%% file: main.tex
\documentclass[10pt,twocolumn,letterpaper]{article}

\usepackage{cvpr}      
\input{preamble}
\definecolor{cvprblue}{rgb}{0.21,0.49,0.74}
\usepackage[pagebackref,breaklinks,colorlinks,allcolors=cvprblue]{hyperref}
\usepackage[accsupp]{axessibility}

\def\paperID{*****} 
\def\confName{CVPR}
\def\confYear{2026}

\title{NTIRE 2026 Challenge on Video Saliency Prediction: Methods and Results}

\author{Andrey Moskalenko~\thanks{A.~Moskalenko (and.v.moskalenko@gmail.com), A.~Bryncev, I.~Kosmynin, K.~Shilovskaya, M.~Erofeev, D.~Vatolin, and R.~Timofte were the challenge organizers, while the other authors participated in the challenge. \cref{affilations} contains the authors’ teams and affiliations. NTIRE 2026 webpage: \url{https://cvlai.net/ntire/2026/}} \and Alexey Bryncev \and Ivan Kosmynin \and Kira Shilovskaya \and Mikhail Erofeev \and Dmitry Vatolin \and Radu Timofte \and Kun Wang \and Yupeng Hu \and Zhiran Li \and Hao Liu \and Qianlong Xiang \and Liqiang Nie \and Konstantinos Chaldaiopoulos \and Niki Efthymiou \and Athanasia Zlatintsi \and Panagiotis Filntisis \and Katerina Pastra \and Petros Maragos \and Li Yang \and Gen Zhan \and Yiting Liao \and Yabin Zhang \and Yuxin Liu \and Xu Wu \and Yunheng Zheng \and Linze Li \and Kun He \and Cong Wu \and Xuefeng Zhu \and Tianyang Xu \and Xiaojun Wu \and Wenzhuo Zhao \and Keren Fu \and Gongyang Li \and Shixiang Shi \and Jianlin Chen \and Haibin Ling \and Yaoxin Jiang \and Guoyi Xu \and Jiajia Liu \and Yaokun Shi \and Jiachen Tu
}

\begin{document}
\maketitle
\input{sec/0_abstract}    
\input{sec/1_intro}
\input{sec/2_challenge}
\input{sec/3_results}
\input{sec/4_solutions}
\input{sec/5_conclusion}
\input{sec/6_affilations}

\section*{Acknowledgments}
The work of Andrey Moskalenko, Alexey Bryncev, and Dmitry Vatolin was financially supported by the Institute for Artificial Intelligence of Lomonosov Moscow State University. The evaluation was carried out using the MSU-270 supercomputer of the Lomonosov Moscow State University. Crowdsourcing assessors were hired using Yandex.Tasks. The submission system was partially supported by the Humboldt Foundation, OPPO, Kuaishou, and the University of Wurzburg (Computer Vision Lab).

{
    \small
    \bibliographystyle{ieeenat_fullname}
    \bibliography{main}
}

\end{document}

%% file: sec/0_abstract.tex
\begin{abstract}
This paper presents an overview of the NTIRE 2026 Challenge on Video Saliency Prediction. The goal of the challenge participants was to develop automatic saliency map prediction methods for the provided video sequences. The novel dataset of 2,000 diverse videos with an open license was prepared for this challenge. The fixations and corresponding saliency maps were collected using crowdsourced mouse tracking and contain viewing data from over 5,000 assessors. Evaluation was performed on a subset of 800 test videos using generally accepted quality metrics. The challenge attracted over 20 teams making submissions, and 7 teams passed the final phase with code review. All data used in this challenge is made publicly available — \url{https://github.com/msu-video-group/NTIRE26_Saliency_Prediction}.
\end{abstract}

%% file: sec/1_intro.tex
\section{Introduction}
\label{sec:intro}

Visual saliency prediction focuses on approximating the behavior of the human visual system (HVS) by modeling how attention is naturally distributed across a scene. Rather than processing all visual information uniformly, humans tend to concentrate on specific regions that stand out perceptually. Saliency prediction models aim to capture and predict these attention patterns automatically. In addition to applications in neuroscience and cognitive science, high-quality saliency estimation is a key component in numerous multimedia applications. It is utilized in tasks such as perceptually guided compression~\cite{hadizadeh2013saliency,gitman_semiautomatic,lyudvichenko2017semiautomatic,patel2021saliency,chang2026saliency}, objective quality evaluation~\cite{zhang2015application,yang2019sgdnet,qu2025kvq,alexey2025bridging}, adaptive content retargeting~\cite{fang2012saliency,ahmadi2021context,guo2024irnet}, media enhancement~\cite{miangoleh2023realistic,safonov2025ntire}, etc. In addition to traditional images and videos, saliency modeling has been extended to immersive formats, including 360° content, where it facilitates optimized encoding and transmission~\cite{wang2022salientvr,wahba2025enhancement,zhao2026rate}. Saliency concepts are also applied in the 3D domain, for example, in the mesh simplification task~\cite{lee2005mesh,dos2023saliency,nousias2023deep}, where geometrical details can be selectively preserved in perceptually important regions.

\subsection{Saliency Prediction Methods}
Early approaches to saliency prediction primarily relied on low-level visual cues, including color, contrast, and texture patterns. These statistical properties of natural scenes formed the foundation of many classical methods~\cite{itti1998model,harel2006graph,judd2009learning}. With the shift toward video analysis, researchers began to incorporate not only spatial but also temporal information. Video saliency prediction models leveraged dynamic features to capture motion across frames~\cite{guo2008spatio,mahadevan2009spatiotemporal,marat2009modelling}.

Modern neural network–based approaches have significantly improved performance by learning complex spatial~\cite{kroner2020contextual,lou2022455}, temporal~\cite{drostejiao2020,jain2021vinet,zhou2023transformer,moradi2024salfom}, and audio~\cite{tavakoli2019dave,chen2022comprehensive,xiong2023casp} patterns directly from data. Newer approaches are also beginning to adapt~\cite{chen2025explainable,tang2025cardiff} large-scale Vision Language Models (VLMs) for saliency prediction tasks. In addition to traditional images and videos, saliency modeling has been extended to other domains, including VR/360° content~\cite{zhang2018saliency,yun2022panoramic,cokelek2025spherical,jiao2026diffgaze} and 3D mesh saliency~\cite{ding2023towards,martin2024sal3d,zhang2025textured,zhang2025mesh}

\subsection{Video Saliency Datasets}
Saliency prediction aims to estimate a continuous saliency map for each frame of an input video. Saliency maps, in turn, have an interpretation of the probability density that a given pixel will attract the user's attention. Saliency maps are typically derived from fixation maps. Fixation maps, in turn, are obtained by aggregating eye-tracking data from a special eye-tracking device on multiple viewers during a viewing session. All fixation maps are then smoothed with a Gaussian, and the resulting map is normalized to correspond to a probability distribution. At present, the largest video datasets containing eye-fixation data are Hollywood-2~\cite{mathe2014actions}, which includes 1707 videos viewed by 19 participants, and DHF1K~\cite{wang2018revisiting}, which contains 1000 videos annotated by 17 viewers.

Collecting eye-tracking data is expensive and difficult to scale, which limits the size of available datasets. To address this, several proxy methods for crowdsourced saliency annotation have been developed. Most of these approaches require only a computer mouse to collect saliency stimuli~\cite{jiang2015salicon,kim2017bubbleview,tavakoli2017saliency,lyudvichenko2019predicting,shaghaghi2025focalvid}, making large-scale crowdsourced data collection feasible. With the aid of various filtering and enhancement techniques, these methods have enabled the creation of large-scale video saliency datasets~\cite{aim_challenge}. In this challenge, we also rely on collecting fixations using mouse movements and have further improved data collection, filtering, and post-processing for obtaining high-quality saliency data in crowdsourcing.

This challenge is one of the challenges associated with the NTIRE 2026 Workshop~\footnote{\url{https://www.cvlai.net/ntire/2026/}} on:
deepfake detection~\cite{ntire26deepfake}, 
high-resolution depth~\cite{ntire26hrdepth},
multi-exposure image fusion~\cite{ntire26raim_fusion}, 
AI flash portrait~\cite{ntire26raim_portrait}, 
professional image quality assessment~\cite{ntire26raim_piqa},
light field super-resolution~\cite{ntire26lightsr},
3D content super-resolution~\cite{ntire263dsr},
bitstream-corrupted video restoration~\cite{ntire26videores},
X-AIGC quality assessment~\cite{ntire26XAIGCqa},
shadow removal~\cite{ntire26shadow},
ambient lighting normalization~\cite{ntire26lightnorm},
controllable Bokeh rendering~\cite{ntire26bokeh},
rip current detection and segmentation~\cite{ntire26ripdetseg},
low light image enhancement~\cite{ntire26llie},
high FPS video frame interpolation~\cite{ntire26highfps},
Night-time dehazing~\cite{ntire26nthaze,ntire26nthaze_rep},
learned ISP with unpaired data~\cite{ntire26isp},
short-form UGC video restoration~\cite{ntire26ugcvideo},
raindrop removal for dual-focused images~\cite{ntire26dual_focus},
image super-resolution (x4)~\cite{ntire26srx4},
photography retouching transfer~\cite{ntire26retouching},
mobile real-world super-resolution~\cite{ntire26rwsr},
remote sensing infrared super-resolution~\cite{ntire26rsirsr},
AI-Generated image detection~\cite{ntire26aigendet},
cross-domain few-shot object detection~\cite{ntire26cdfsod},
financial receipt restoration and reasoning~\cite{ntire26finrec},
real-world face restoration~\cite{ntire26faceres},
reflection removal~\cite{ntire26reflection},
anomaly detection of face enhancement~\cite{ntire26anomalydet},
video saliency prediction~\cite{ntire26videosal},
efficient super-resolution~\cite{ntire26effsr},
3d restoration and reconstruction in adverse conditions~\cite{ntire26realx3d},
image denoising~\cite{ntire26denoising},
blind computational aberration correction~\cite{ntire26aberration},
event-based image deblurring~\cite{ntire26eventblurr},
efficient burst HDR and restoration~\cite{ntire26bursthdr},
low-light enhancement: `twilight cowboy'~\cite{ntire26twilight},
and efficient low light image enhancement~\cite{ntire26effllie}.

%% file: sec/2_challenge.tex
\section{Video Saliency Prediction Challenge}
\label{sec:challenge}

\subsection{Challenge Dataset}

We used two publicly available datasets as sources of video content~\cite{wang2024youtube,Farre2024FineVideo}. Both datasets consist of diverse videos curated from YouTube, and all included videos are distributed under an open CC-BY license.

As an additional preprocessing step, we retained only video streams with a minimum resolution of FullHD. All selected videos were then transcoded using the libx264 codec with a constant rate factor (CRF) of 23, converted to 30 FPS, and resized to FullHD resolution. Although all videos shared the same aspect ratio, they could be presented in either horizontal or vertical orientation. Horizontal and vertical videos were sampled at a 3:1 ratio. In addition, audio tracks were normalized according to the EBU R128 standard and transcoded to stereo AAC at 256 Kbps.

We built on prior work in crowdsourced saliency data collection, including our previous methodology~\cite{aim_challenge}, and further refined the procedure through participant filtering as well as additional pre- and post-processing of fixation data. These improvements brought the quality of the collected data closer to that obtained with an eye-tracker.

The obtained challenge dataset consists of 2,000 videos and over 1M frames. Over 5,000 viewers participated in the crowdsourcing data collection, resulting \textgreater70 viewers for each video. The average assessor time it takes to complete the 20 main and 3 verification questions is 15 minutes. The average video length in the dataset is 18 seconds.

\subsection{Evaluation}
\label{evaluation}
To objectively evaluate challenge solutions, we employed four widely used metrics~\cite{bylinskii2018different}: Pearson’s Correlation Coefficient (CC), Similarity (SIM), Area Under the Curve (AUC-Judd), and Normalized Scanpath Saliency (NSS). For each participant, the final ranking was determined by averaging their ranks across all four metrics on the test set. In cases where two or more methods obtained the same final rank, ties were resolved using the first metric in the listed order for which their scores differed. During the final phase, participants were required to submit their final predictions, a factsheet, and the code necessary to reproduce the reported results; these submissions were subsequently verified by the organizers. In total, 7 teams successfully completed this phase. The final evaluation also included the organizers’ baseline Center Prior method.

\begin{table*}
\fontsize{10pt}{13.25pt}\selectfont
\tabcolsep=4pt
\begin{tabular}{clcccclccccccc}
\toprule
\multicolumn{1}{l}{} &  & \multicolumn{4}{c}{Public Test Subset Metrics}                        & \multicolumn{1}{c}{} & \multicolumn{4}{c}{Private Test Subset Metrics}                       &                      & \multicolumn{2}{c}{Additional info}      \\
Team Name            &  & CC              & SIM             & AUC-Judd        & NSS             &                      & CC              & SIM             & AUC-Judd        & NSS             &                      & Rank & \#Params(M)          \\ \cline{1-1} \cline{3-6} \cline{8-11} \cline{13-14} 
iLearn               &  & \underline{0.8369}    & \textbf{0.6989} & \underline{0.8970}    & \textit{3.4381}          &                      & \textbf{0.8280} & \textbf{0.6927} & \underline{0.8921}    & \textit{3.3229}          &                      & 1.75              &       6880               \\
CVSP                 &  & \textbf{0.8374} & \underline{0.6723}    & \textbf{0.9032} & \underline{3.5440}    &                      & \underline{0.8272}    & \underline{0.6640}    & \textbf{0.8982} & \underline{3.4156}    &                      & 1.75              &   4229                   \\
ARK\_MMLAB            &  & 0.7977          & 0.6636          & \textit{0.8961}          & \textbf{3.5963} &                      & 0.7896          & 0.6597          & \textit{0.8913}          & \textbf{3.4562} &                      & 3.00                 &          2216            \\
Vertex               &  & \textit{0.8071}          & \textit{0.6700}          & 0.8910          & 3.2828          &                      & \textit{0.7962}          & \textit{0.6636}          & 0.8856          & 3.1543          &                      & 3.50               &          534            \\
AAM                  &  & 0.7546          & 0.5837          & 0.8181          & 3.2073          &                      & 0.7484          & 0.5790          & 0.8145          & 3.0996          &                      & 5.75              &  425 \\
SHU-MIIPLab          &  & 0.7228          & 0.5920          & 0.8850          & 2.7915          &                      & 0.7174          & 0.5928          & 0.8797          & 2.7014          &                      & 5.75              &         860             \\
NTR                  &  & 0.7028          & 0.5849          & 0.8519          & 3.1231          &                      & 0.6927          & 0.5788          & 0.8452          & 2.9893          & \multicolumn{1}{l}{} & 6.50               &         64            \\
Baseline             &  & 0.4175          & 0.4086          & 0.6968          & 1.3709          &                      & 0.4101          & 0.4081          & 0.6911          & 1.3054          &                      & 8.00                 & —          \\ \bottomrule         
\end{tabular}
\caption{Results of NTIRE 2026 Video Saliency Prediction Challenge. Best scores are shown in \textbf{bold}, the second best is \underline{underlined}, while the third best is \textit{italic}. The ranking is based on the mean rank across all the metrics (\cref{evaluation}) on Private Test Subset. In cases where two or more methods obtained the same final rank, ties were resolved using the first metric in the listed order for which their scores differed. The \#Params column describes for each model the number of parameters in millions.}
\label{table:results}
\end{table*}

\subsection{Challenge Phases}
The challenge dataset was randomly divided into two subsets: a training set comprising 1,200 videos with corresponding fixation data and saliency maps provided to participants, and a test set containing 800 videos. The test set was further randomly split into a public test subset of 300 videos, used for online evaluation during the competition, and a private test subset of 500 videos. Throughout the competition, participants could submit their predictions and view performance metrics on the public test subset. During the two intermediate phases, each participant’s best-performing solution on the public subset was evaluated on 250 videos from the private test subset. In the final phase, all methods were assessed on the entire private test subset.

%% file: sec/3_results.tex
\section{Results}
This section introduces the results of NTIRE 2026 Video Saliency Prediction Challenge. The values of all metrics and the final ranking are presented in Table~\ref{table:results}. In this case, two teams have the same average rank across metrics, and according to the metric priority rules, the team with the highest CC metric value wins. 

The first-place solution received from the iLearn team (\cref{ilearn}). They used a shared InternVideo2~\cite{internvideo2} backbone with two complementary decoders, and the final saliency map is produced by averaging the two expert predictions. The second-place solution received from the CVSP team (\cref{cvsp}). They used a large-scale V-JEPA2~\cite{assran2025v} backbone, using its self-supervised video representations to estimate saliency from predictive spatiotemporal features. The third-place solution from ARK\_MMLAB (\cref{ark_mmlab}) builds a hierarchical model on InternVideo2~\cite{internvideo2}, extracting multi-level spatiotemporal features and aligning them with a feature upsampling module. Vertex team~\ref{vertex} extends TMFI~\cite{zhou2023transformer} by adding a bottom-up aggregation path to the original top-down fusion, giving richer multi-scale feature interaction. AAM team~\ref{aam} combines audio-visual fusion, temporal dynamics modeling, and a hyperbolic decoder to predict saliency maps. SHU-MIIPLab team~\ref{shumiiplab} proposes a diffusion-based model that predicts saliency from RGB frames and optical flow using state-space blocks. NTR team~\ref{ntr} uses a dual-stream design, combining a pretrained video backbone for motion cues with a pretrained image backbone for fine spatial details.

%% file: sec/4_solutions.tex
\section{Teams and Methods}

\subsection{iLearn}
\label{ilearn}

\begin{figure}[ht]
    \centering
    \includegraphics[width=1.0\columnwidth]{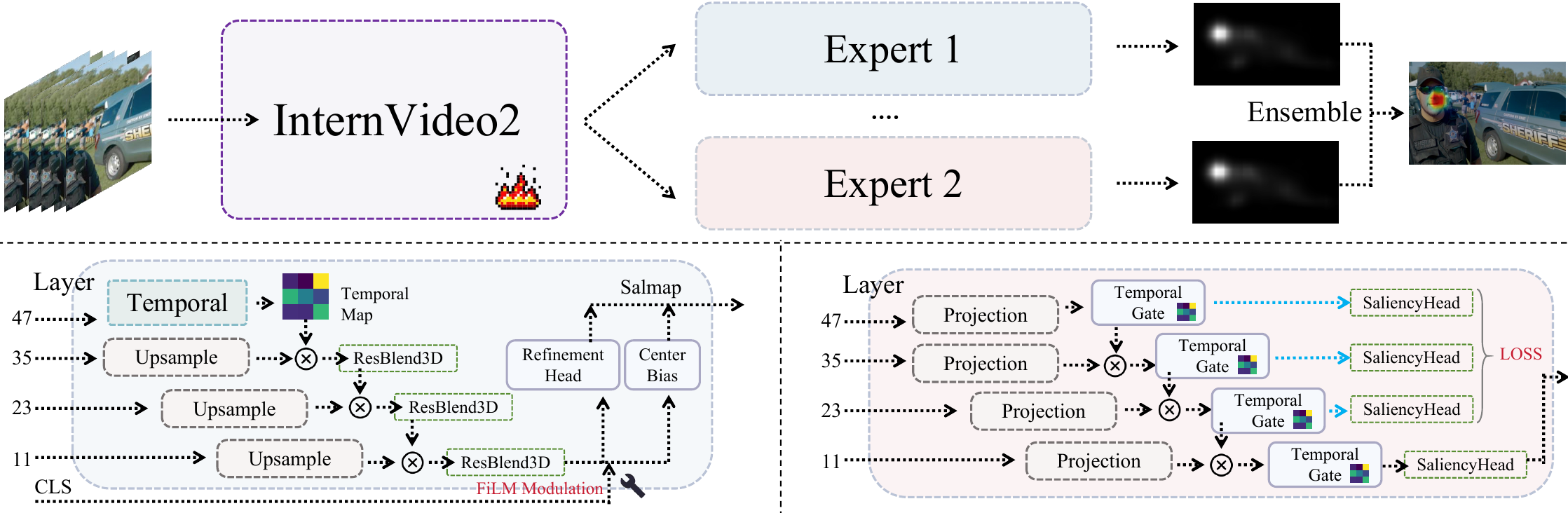}
    \caption{iLearn Video Saliency Prediction Pipeline.}
    \label{fig:team_name}
\end{figure}

We propose a multi-expert ensemble framework~\cite{ntire26visage} for video saliency prediction. As illustrated in Figure~\ref{fig:team_name}, our method builds upon a shared InternVideo2 \cite{internvideo2} backbone and employs two structurally complementary decoders whose predictions are fused to produce the final saliency maps.

\textbf{Shared Backbone.} We adopt InternVideo2-Stage2\_6B as the spatio-temporal feature extractor. Multi-level features are extracted from four intermediate layers (layers 11, 23, 35, and 47), capturing information ranging from low-level spatial details to high-level semantic abstractions. The backbone is fine-tuned with LoRA \cite{lora} for saliency prediction, while preserving the pretrained video representations.

\textbf{Expert 1: Temporal-Modulated Decoder with Spatial Priors.} The first decoder generates an explicit temporal attention map from the deepest features (layer 47) and uses it to multiplicatively modulate shallower features in a top-down manner. At each level, the modulated features are refined through 3D residual blending blocks. Feature-wise Linear Modulation (FiLM) \cite{film} is further applied to inject global conditioning. The output is passed through a refinement head augmented with a learnable center bias prior, leveraging the well-known tendency of human gaze toward the image center.

\textbf{Expert 2: Multi-Scale Decoder with Deep Auxiliary Supervision.} The second decoder projects all four levels of features into a unified dimension and fuses them in a top-down method via concatenation and 3D residual fusion blocks. A temporal gating mechanism is applied at the deepest level to re-weight features across time steps. Crucially, this decoder employs deep auxiliary supervision: independent saliency prediction heads are attached at multiple intermediate levels, providing dense gradient signals that encourage each level to learn meaningful saliency representations.

\textbf{Complementarity and Ensemble.} The two experts offer complementary inductive biases. Expert 1 relies on multiplicative gating and explicit spatial priors, excelling at scenes with dominant central subjects and stable temporal dynamics. Expert 2 employs concatenation-based fusion and data-driven multi-scale learning without built-in priors, offering stronger adaptability to complex scenes with off-center or multiple salient regions. To fuse the complementary predictions, we adopt a logit-space ensemble. The saliency outputs from both experts are first transformed into logits via the inverse sigmoid function, averaged, and then projected back to the probability space using a sigmoid activation. Furthermore, our framework is extensible; while the current implementation focuses on these two experts, additional experts tailored to specific requirements can be integrated into the ensemble in the future.

\textbf{Training Strategy.} We adopt a two-stage training scheme. In Stage 1, the backbone is frozen and only the decoder is trained to establish a strong saliency decoding capability. In Stage 2, LoRA modules are injected into the backbone and jointly fine-tuned with the decoder using a lower learning rate. The training objective combines KL divergence, Pearson's correlation coefficient, similarity, and normalized scanpath saliency losses. Each expert is trained independently following this two-stage protocol, and the ensemble is performed at inference time.

\begin{figure}[!ht]
  \centering
  \includegraphics[width=\linewidth]{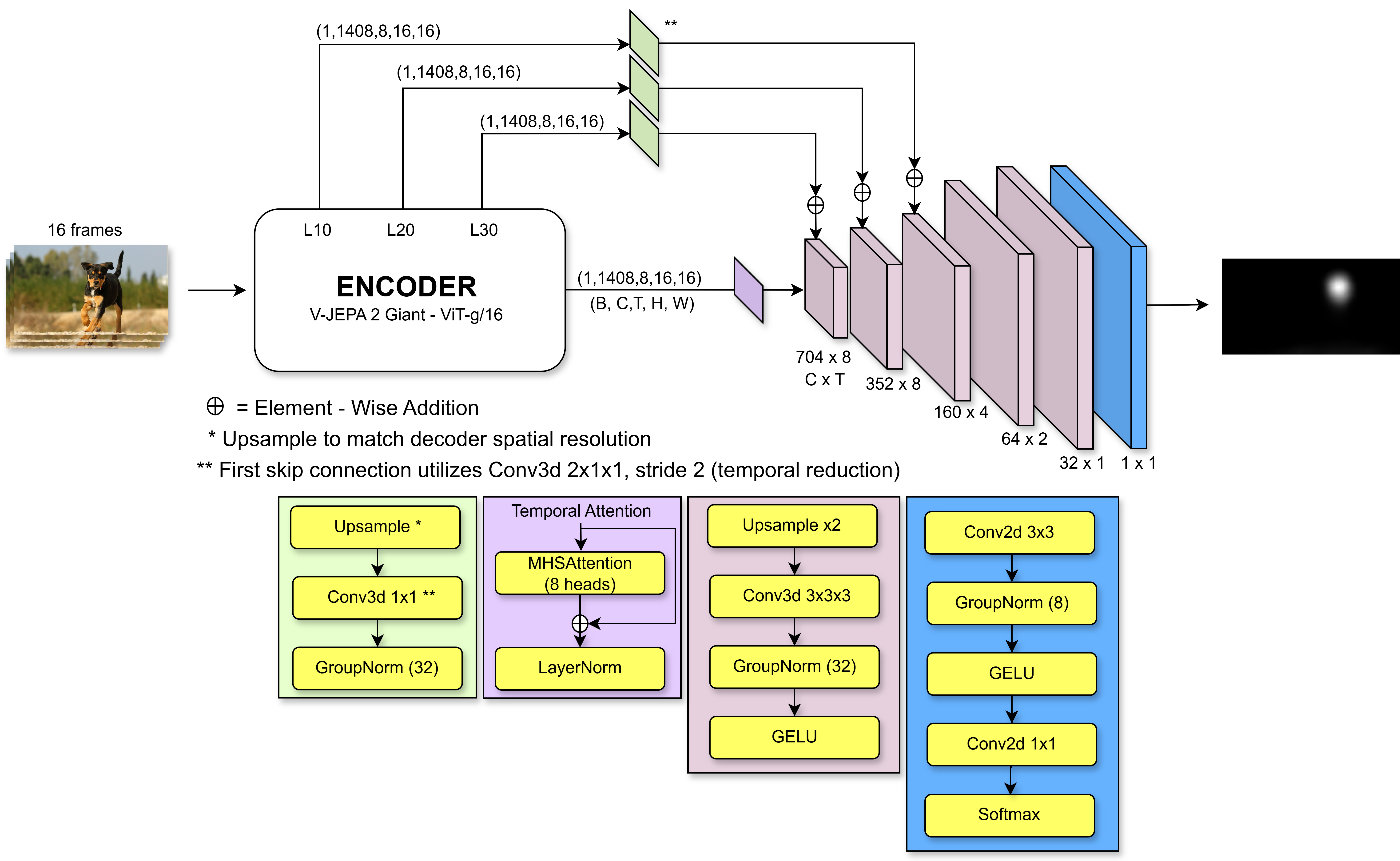}
  \caption{An overview of PredJSal. The framework repurposes V-JEPA2 backbone to extract rich spatiotemporal representations, decoded into per-frame saliency maps via a 3D convolutional decoder with multi-scale skip connections and a lightweight temporal self-attention module.}
  \label{fig:arch}
\end{figure}

\subsection{CVSP}
\label{cvsp}
Human visual attention is strongly guided by an implicit understanding of how the physical world behaves. Our gaze naturally gravitates towards objects that move unexpectedly or violate our expectations about a scene.
Predictive coding theory~\cite{rao1999predictive} interprets this behavior, proposing that the brain minimizes prediction errors relative to a generative model of the world. Under this framework, prediction errors drive action to reduce them, and saliency, as a driver of attention and eye movements, should therefore correlate with the prediction errors produced by a predictive coding model~\cite{spratling2012}.

This theoretical foundation motivates our use of V-JEPA2~\cite{assran2025v} as a backbone for saliency prediction, since it is trained to predict masked regions of natural videos in a learned representation space. V-JEPA2 acquires an implicit understanding of intuitive physics, including object permanence and shape consistency~\cite{garrido2025intuitive}, through the same principle of minimizing spatiotemporal prediction error, making its representations a substrate for saliency estimation.

We therefore present PredJSal~\cite{chaldaiopoulos2026predjsal}, a predictive JEPA-based framework that repurposes the rich internal representations of V-JEPA2~\cite{assran2025v} for video saliency prediction. The overall pipeline is shown in Figure~\ref{fig:arch}. 
PredJSal is built on a V-JEPA2 Giant (ViT-g/16) backbone pretrained with self-supervised masked video prediction on large-scale video data. Given a clip of frames, the encoder produces a dense spatiotemporal feature volume $\mathbf{F}$, with intermediate representations extracted at three evenly spaced encoder layers to provide coarse-to-fine features for multi-scale decoding. Before decoding, a lightweight self-attention module refines the encoder features along the temporal axis, treating each spatial position as an independent sequence over time to integrate cross-frame context without mixing spatial locations, producing $\hat{\mathbf{F}}$. 

A 3D convolutional decoder then progressively upsamples $\hat{\mathbf{F}}$ while reducing the temporal dimension to a single output frame, with multi-scale skip connections from the intermediate encoder layers fused via element-wise addition at three decoder stages to recover spatial detail lost during encoding. A lightweight 2D convolutional head produces our final prediction.

To reduce the effect of random initialization, we train our model on four different seeds and average the predicted saliency maps at inference to produce a single map.

\begin{figure}[!ht]
    \centering
    \includegraphics[width=\linewidth]{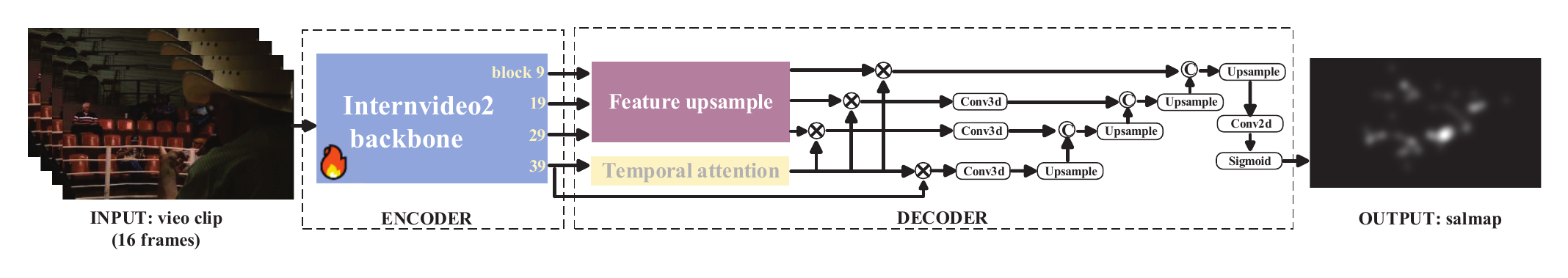}
    \caption{ARK\_MMLAB video saliency prediction framework.}
    \label{fig:pipeline}
\end{figure}

\subsection{ARK\_MMLAB}
\label{ark_mmlab}
Our proposed video saliency prediction framework is built upon a hierarchical architecture, aiming to effectively capture both spatial semantics and temporal dynamics from video clips. As illustrated in Figure~\ref{fig:pipeline}, the pipeline consists of a visual encoder, a feature upsampling module, a temporal attention mechanism, and a hierarchical decoder. The detailed structures of our method are illustrated as follows.

\subsubsection{Visual Encoder}
We adopt the \textbf{InternVideo2-1B} \cite{wang2024internvideo2} model as our backbone to extract robust spatiotemporal representations. Given an input video clip of size $3 \times 16 \times 224 \times 224$ (Channel $\times$ Time $\times$ Height $\times$ Width), the encoder generates a hierarchy of feature maps from four different depth levels, denoted as $y_1, y_2, y_3$, and $y_4$. These features encapsulate fine-grained local details at early stages and abstract high-level semantics at deeper stages. To align the dimensions for subsequent processing, all extracted features are reshaped into 5D tensors of size $1408 \times 16 \times H \times W$, where the spatial resolutions are $112 \times 112$, $56 \times 56$, $28 \times 28$, and $14 \times 14$.

\subsubsection{Feature Upsampling Module}
To mitigate the spatial resolution discrepancy between different feature levels, we introduce a \textbf{Feature Upsampling Module}. This module applies a series of bilinear interpolations combined with 2D Convolutions, Group Normalization, and GELU activations. Specifically, $y_1$ is upsampled to a spatial size of $128 \times 128$ with $512$ channels, $y_2$ to $64 \times 64$ with $512$ channels, and $y_3$ to $32 \times 32$ with $768$ channels. The deepest feature $y_4$ is retained at its original resolution ($14 \times 14$) to preserve the highest level of semantic abstraction.

\subsubsection{Temporal Attention and Hierarchical Decoder}
Before decoding, a \textbf{Temporal Attention} mechanism is applied. It takes the deepest feature $y_4$ to compute a set of temporal weights using 3D convolutions and temporal pooling. These weights are then multiplied across all feature levels ($y_1$ to $y_4$) to dynamically emphasize the most salient frames within the clip.

\begin{figure}[ht]
    \centering
    \includegraphics[width=\linewidth]{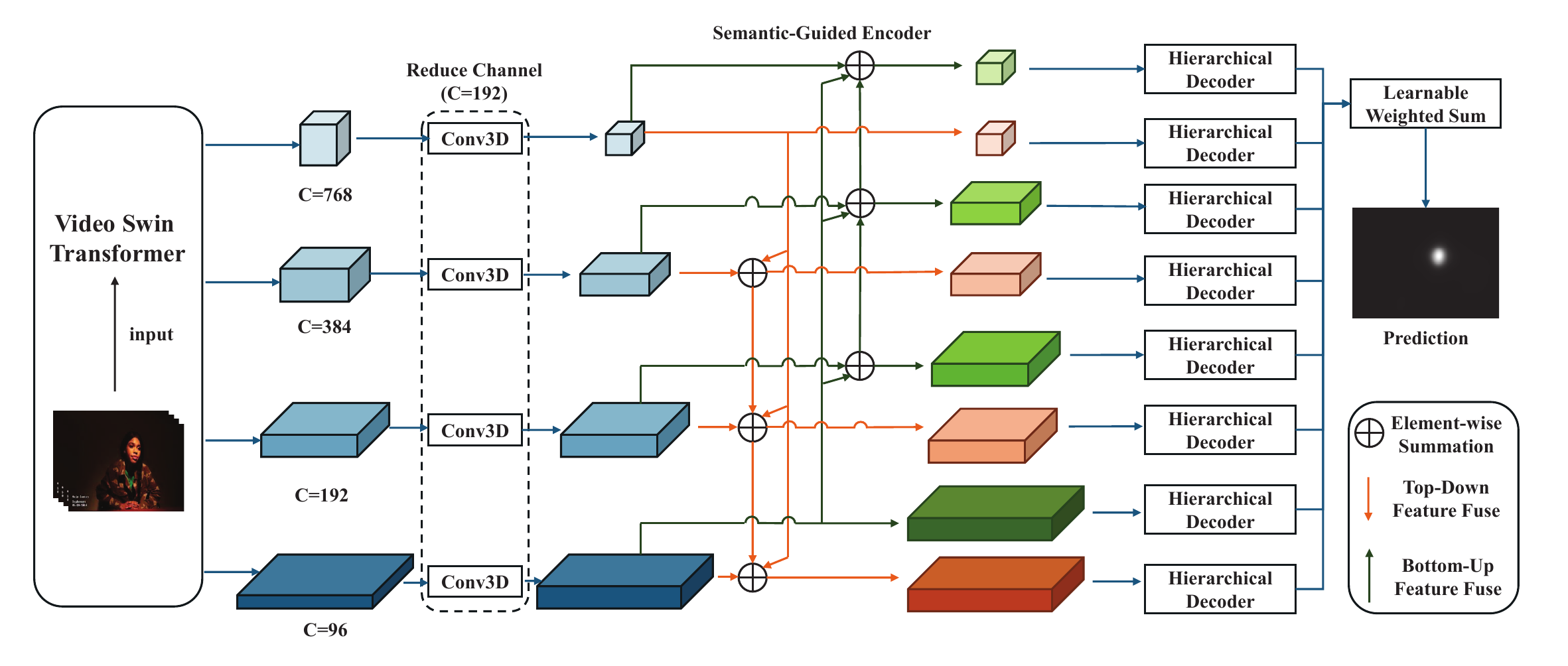}
    \caption{Pipeline of the proposed bidirectional enhanced TMFI.
    We extend TMFI by introducing an additional bottom-up multi-scale feature fusion to more comprehensively exploit multi-scale information. In the figure, the orange arrows represent the original top-down fusion in TMFI, while the green arrows denote the added bottom-up fusion. During both fusion processes, the highest-level or lowest-level features are repeatedly utilized, as illustrated by the diagonal arrows in the corresponding colors.}
    \label{fig:vertex_pipeline}
\end{figure}

The \textbf{Hierarchical Decoder} progressively fuses these temporally-weighted features in a top-down manner. It consists of four branches corresponding to the four feature levels. Starting from the deepest branch (Branch 4), the features are processed through 3D Convolutions and ReLU activations, upsampled, and then concatenated with the features from the adjacent shallower branch (Branch 3). This concatenation and upsampling process is repeated until Branch 1. Finally, the fused representations are passed through a bilinear interpolation layer and a series of 2D convolutions to yield the final saliency map of size $224 \times 224$.

\subsection{Vertex}
\label{vertex}
TMFI~\cite{zhou2023transformer} is one of the state-of-the-art methods for video saliency prediction. It employs a Video Swin Transformer~\cite{liu2022video} to extract four-scale spatiotemporal features, followed by a Semantic-Guided Encoder that enhances these features in a top-down manner. The enhanced multi-scale features are then fed into a Hierarchical Decoder to produce corresponding saliency maps, which are fused via a 3D convolution to obtain the final prediction.

Building upon TMFI, we extend the original top-down enhancement by introducing an additional bottom-up feature aggregation pathway, as illustrated in Figure~\ref{fig:vertex_pipeline}. This bidirectional design enables better utilization of both high-level semantic information and low-level fine-grained details. Specifically, while TMFI adopts upsampling for top-down fusion, we perform bottom-up aggregation by applying max-pooling to lower-level features for downsampling to align feature resolutions. The aligned features are then combined via element-wise addition and further refined using 3D convolutions for deeper spatiotemporal fusion. As a result, the number of feature scales is expanded from 4 to 8, leading to richer representations. Correspondingly, we generate 8 multi-scale saliency maps, which are subsequently fused into the final prediction via a learnable weighted aggregation approach, replacing the original 3D convolution for more efficient and stable fusion.

\begin{figure}[ht]
    \centering
    \includegraphics[width=\linewidth]{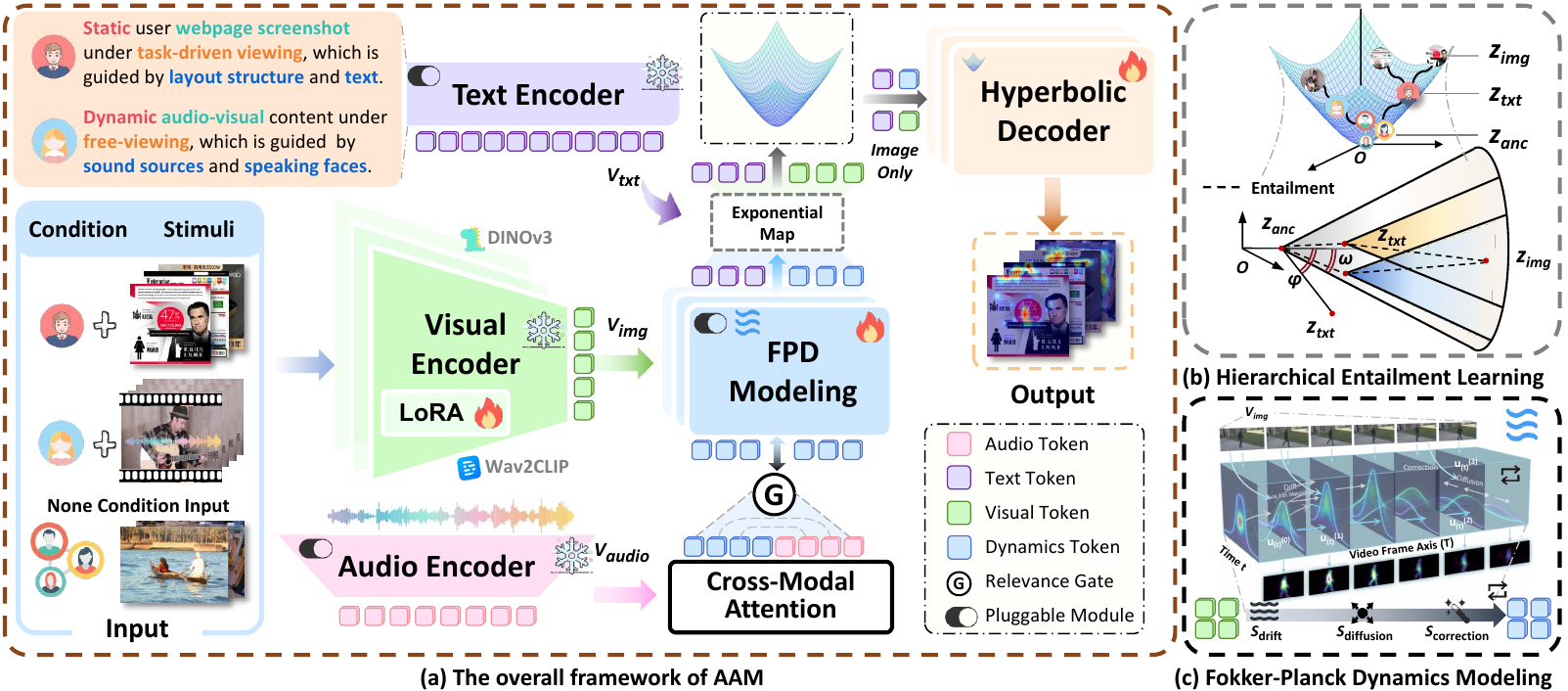}
    \caption{(a) Overview of AAM. (b) The general-specific attention ordering between ($\mathbf z_{\mathrm{anc}}$,$\mathbf z_{\mathrm{txt}}$), ($\mathbf z_{\mathrm{txt}}$,$\mathbf z_{\mathrm{img}}$) is enforced in hyperbolic space using entailment cones. The external angle $\phi$ of a specific condition ($\mathbf z_{\mathrm{txt}}$) is pushed to be within the aperture threshold $\eta\omega$ of the general attention ($\mathbf z_{\mathrm{anc}}$).
    (c) The Fokker--Planck Dynamics (FPD) Modeling illustrates the drift, diffusion, and correction processes used to model the transition of attention over the video frame axis.}
    \label{fig_overview}
\end{figure}

Furthermore, we observe strong complementarity across different model paradigms. Therefore, we incorporate additional predictions from DiffSal~\cite{xiong2024diffsal}, SalFoM~\cite{moradi2024salfom}, and the original TMFI, and perform a unified fusion strategy. The final prediction is obtained through a weighted combination of outputs from all these models, followed by explicit post-processing operations including Gaussian smoothing and temporal smoothing, which improve spatial coherence and reduce temporal inconsistencies across consecutive frames.

\subsection{AAM}
\label{aam}
We present the Attend to Anything Model (AAM), which represents human attention as a shared latent process across modalities, scenes, and time by modeling hierarchical semantic specialization and temporal dynamics, as shown in Figure~\ref{fig_overview}. \textbf{Visual} inputs are encoded by a frozen self-supervised backbone (DINOv3~\cite{simeoni2025dinov3}) with LoRA for adaptation to attention modeling.
\textbf{Text} prompts and \textbf{audio} signals are encoded using frozen CLIP~\cite{radford2021learning} and Wav2CLIP~\cite{wu2022wav2clip} encoders, respectively, with audio mapped into the visual semantic space.
Audio-visual fusion is performed through a relevance-gated cross-attention mechanism, ensuring that audio cues contribute only when semantically aligned.
For video inputs, frame-wise attention representations are refined by a Fokker--Planck Dynamics (FPD) module that models attention evolution over a spatiotemporal manifold.
Visual and textual representations are lifted into hyperbolic space via hierarchical entailment learning for explicit hierarchy enforcement.
Finally, a geometry-aware hyperbolic decoder projects structured representations back to Euclidean space to generate spatial attention maps.

\begin{figure}[!ht]
    \centering
    \includegraphics[width=1.0\columnwidth]{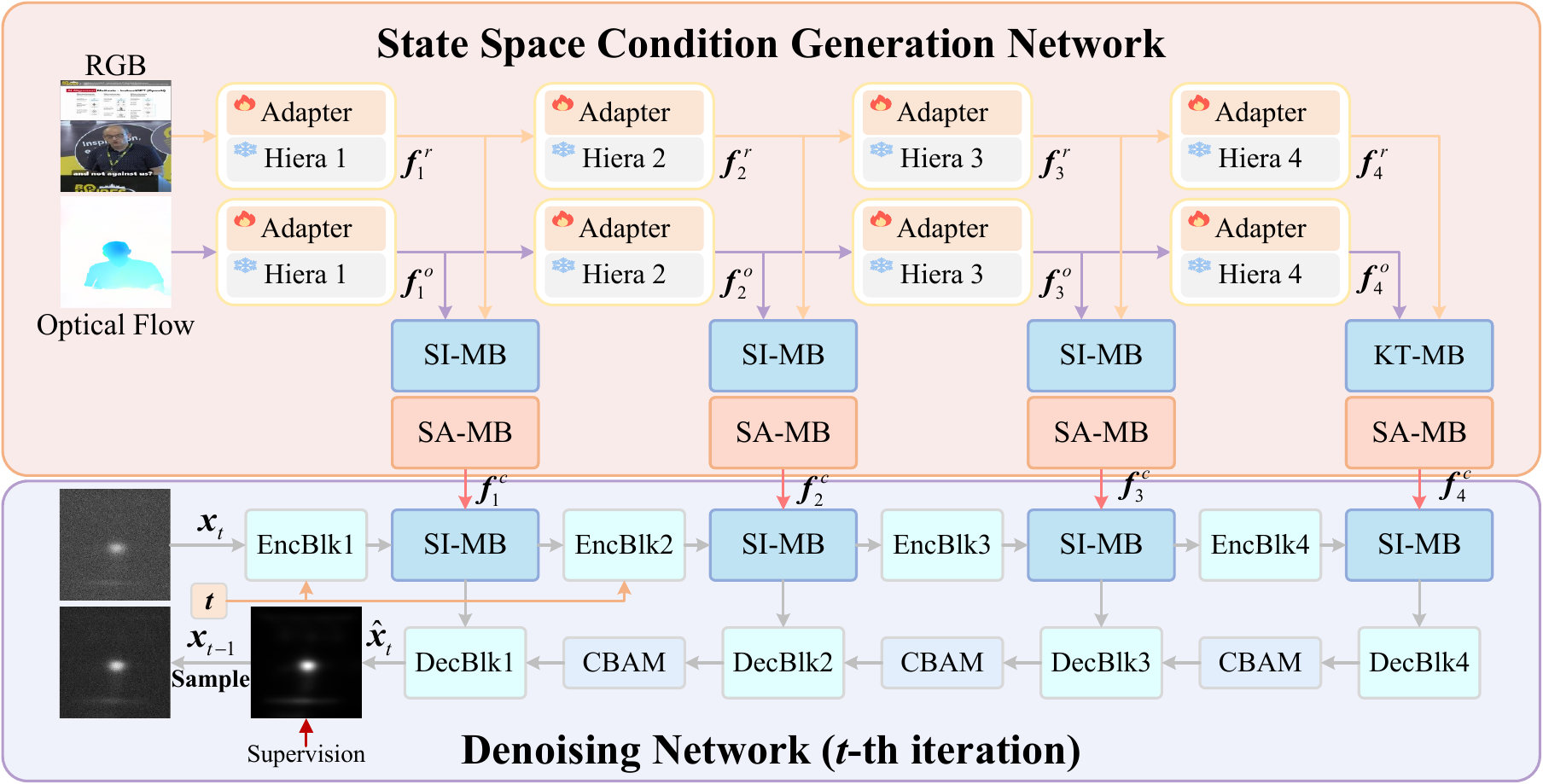}
    \caption{Proposed SSF-DiffNet pipeline, which consists of a state space condition generation network and a denoising network.}
    \label{fig:shu-miiplab_pipeline}
\end{figure}

\subsection{SHU-MIIPLab}
\label{shumiiplab}
We propose \textbf{SSF-DiffNet}, a novel \textbf{S}tate \textbf{S}pace \textbf{F}usion-driven \textbf{Diff}usion \textbf{Net}work for video saliency prediction.
SSF-DiffNet is a denoising diffusion model~\cite{ho2020ddpm,rombach2022latent}.
The inputs are an RGB frame and its corresponding optical flow map generated by Flow-Anything~\cite{liang2025flowanything}.

As illustrated in Figure~\ref{fig:shu-miiplab_pipeline}, SSF-DiffNet consists of two major components: a denoising network tailored for generating saliency maps and a state space condition generation network that works with the denoising network.
The state space condition generation network is constructed based on the state space model~\cite{gu2023mamba,liu2024vmamba}.

Specifically, we first adopt the SAM2 encoder~\cite{ravi2025sam2} (i.e.\ Hiera~\cite{ryali2023hiera}) equipped with adapters as the feature extraction backbone for RGB and optical flow. Then, we utilize RGB to guide optical flow through a Semantic Injection Mamba Block (SI-MB), thereby transmitting semantic attributes to motion cues to identify salient regions. Subsequently, we integrate RGB with optical flow through a Spatiotemporal Alignment Mamba Block (SA-MB). The SA-MB aligns the appearance and motion features and narrows their discrepancies, generating high-quality conditions to guide the denoising network. In the denoising network, we reuse SI-MB to transmit texture cues and semantic priors from these highly informative conditions to the noisy mask, producing accurate saliency maps through iterative optimization. With all components collaborating tightly, our SSF-DiffNet achieves competitive performance.

\subsection{NTR}
\label{ntr}
\subsubsection{Dual-Stream Network Architecture}

We propose a \textbf{Pretrained Dual-Stream Saliency Network} that jointly exploits spatiotemporal motion cues from a video stream and fine-grained spatial appearance cues from a frame stream. Both streams are initialized from domain-specific pretrained weights, enabling the network to leverage large-scale video and image priors without requiring additional pretraining on saliency data.

\subsubsection{Video Stream}

The video stream processes a clip of $L{=}8$ frames as input. We use R(2+1)D-18~\cite{tran2018r21d}, a factored spatiotemporal convolutional network pretrained on the Kinetics-400~\cite{kay2017kinetics} action recognition dataset. The (2+1)D factorization separates 3D convolutions into spatial 2D and temporal 1D components, improving both training efficiency and motion modeling.

Given an $L$-frame clip $\mathbf{C} \in \mathbb{R}^{L \times 3 \times H' \times W'}$, the R(2+1)D backbone produces feature maps at four spatial scales: $\{V^{(1)}, V^{(2)}, V^{(3)}, V^{(4)}\} = \text{R(2+1)D-18}(\mathbf{C})$,
where $V^{(s)} \in \mathbb{R}^{T_s \times C_s \times H_s \times W_s}$ has a temporal dimension $T_s$ that decreases with depth. To obtain spatial feature maps, we apply \emph{temporal softmax pooling}: the temporal logits are computed as a learned linear projection of the channel dimension, softmax-normalized across the $T_s$ frames, and used to compute a weighted sum: $ \bar{V}^{(s)} = \sum_{j=1}^{T_s} \alpha_j^{(s)} V_j^{(s)}, \quad {\alpha}^{(s)} = \text{softmax}(\mathbf{w}^{(s)} \cdot V^{(s)})$,
yielding temporally aggregated features $\bar{V}^{(s)} \in \mathbb{R}^{C_s \times H_s \times W_s}$.

\subsubsection{Frame Stream}

The frame stream processes only the center frame $I_t$ to capture high-resolution spatial appearance. We use ConvNeXt-Tiny~\cite{liu2022convnext}, pretrained on ImageNet-1K~\cite{deng2009imagenet}, which provides 4 feature stages: $\{P^{(1)}, P^{(2)}, P^{(3)}, P^{(4)}\} = \text{ConvNeXt-Tiny}(I_t)$,
where $P^{(s)} \in \mathbb{R}^{C'_s \times H_s \times W_s}$. The ConvNeXt features provide complementary static-scene context that complements the motion-focused R(2+1)D features.

\subsubsection{Decoder with Center Prior Fusion}

The decoder merges the two streams across all four scales via learned $1{\times}1$ projections followed by concatenation and $3{\times}3$ convolutions. Starting from the coarsest scale and progressing to the finest, each decoder stage fuses: $D^{(s)} = \text{Conv}\bigl([\bar{V}^{(s)}, P^{(s)}, \text{Up}(D^{(s+1)})]\bigr)$,
where $[\cdot]$ denotes channel-wise concatenation and $\text{Up}(\cdot)$ is bilinear upsampling. At the decoder output, we fuse a Gaussian \textbf{center prior} $G \in [0,1]^{H' \times W'}$~\cite{judd2009learning}. The center prior is a precomputed 2D Gaussian that captures the well-known center bias of human gaze. The final saliency logit is: $\hat{S}_t = \sigma\bigl(D^{(1)} + \beta \cdot G\bigr)$,
where $\sigma$ is the sigmoid function and $\beta$ is a learned scalar weight. The output is bilinearly upsampled to the original video resolution.

\subsubsection{Loss Function}

We combine four complementary losses: $\mathcal{L} = \lambda_{\text{KL}}\,\mathcal{L}_{\text{KL}} + \lambda_{\text{CC}}\,\mathcal{L}_{\text{CC}} + \lambda_{\text{NSS}}\,\mathcal{L}_{\text{NSS}} + \lambda_{\text{BCE}}\,\mathcal{L}_{\text{BCE}}$, with $\lambda_{\text{KL}}{=}1.0$, $\lambda_{\text{CC}}{=}0.7$, $\lambda_{\text{NSS}}{=}0.3$, $\lambda_{\text{BCE}}{=}0.1$. \textbf{KL divergence} $\mathcal{L}_{\text{KL}}$ measures the information loss between predicted and ground-truth saliency distributions. \textbf{Correlation coefficient} $\mathcal{L}_{\text{CC}} = 1 - \text{CC}(\hat{S}_t, S_t)$ directly penalizes low correlation. \textbf{NSS} $\mathcal{L}_{\text{NSS}} = -\text{NSS}(\hat{S}_t, F_t)$ rewards high normalized scanpath saliency at fixation locations. \textbf{Fixation BCE} $\mathcal{L}_{\text{BCE}}$ applies binary cross-entropy with positive class weight 8.0 to the binary fixation map $F_t$, sharpening saliency at fixation sites.

\begin{table}[ht]
\centering
\small
\caption{Training configuration.}
\begin{tabular}{@{}lcc@{}}
\toprule
\textbf{Parameter} & \textbf{Stage 1} & \textbf{Stage 2} \\
\midrule
Init from & ImageNet/K400 & Stage 1 best \\
Epochs & 4 & 3 \\
Batch size & 8 & 8 \\
Input size & $224^2$ & $224^2$ \\
Clip length & 8 frames & 8 frames \\
Frame stride & 2 & 2 \\
Clips/video & 4 & 4 \\
LR (head) & $2\!\times\!10^{-4}$ & $1\!\times\!10^{-4}$ \\
LR (backbone) & $2\!\times\!10^{-5}$ & $1\!\times\!10^{-5}$ \\
Weight decay & $10^{-4}$ & $10^{-4}$ \\
Grad clip & 1.0 & 1.0 \\
Train/Val split & 1100/100 & 1100/100 \\
\bottomrule
\end{tabular}
\label{tab:training}
\end{table}

\subsubsection{Warm-Start Training Schedule}

Training proceeds in two stages, each warm-starting from the previous best checkpoint. Both stages use AdamW with cosine annealing. The backbone learning rate is set to $0.1\times$ the head learning rate to avoid disrupting the pretrained spatiotemporal and image features.

\subsubsection{Inference}
Each output frame is predicted using an 8-frame clip ($L{=}8$, stride 2) centered on that frame, processed at $224{\times}224$ resolution. The batch size is 8 clips per forward pass. 

\textbf{Test-time augmentation (TTA):} We apply horizontal flip TTA — each clip is processed in its original orientation and horizontally flipped, then the two predicted saliency maps are averaged after flipping the second back to the canonical orientation. This halves variance from left-right saliency asymmetries. \textbf{Output:} Per-frame saliency maps are bilinearly upsampled to the original video resolution, min-max normalized per video, and written as grayscale MP4 (all 3 BGR channels equal) at the original FPS.

\subsection{Baseline Solution}
This solution was provided by the organizers' team as a baseline. To obtain the Center Prior prediction, we averaged all saliency maps from all 1,200 training videos, and then fitted a Gaussian function centered at the center of the frame, optimizing only the $\sigma_{x}, \sigma_{y}$ parameters. The fitting target was $L2$ loss between the two images. The resulting Center Prior was duplicated for all frames for all test videos and given to the participants as a sample submission.

%% file: sec/5_conclusion.tex
\section{Conclusion}
This paper presented an overview of the NTIRE 2026 Challenge on Video Saliency Prediction. We introduced a novel dataset of 2,000 videos, accompanied by saliency annotations collected from more than 5,000 assessors through crowdsourced mouse tracking. In the final phase, 7 teams proposed their solutions and passed the code-sharing phase. The final results show that top solutions are built on a large pretrained video backbones and multi-scale spatiotemporal modeling. We believe that the described methods and challenge dataset will support future research toward more accurate video saliency prediction models.

%% file: sec/6_affilations.tex
\section{Teams and Affiliations}\label{affilations}
\label{sec:aff}

\subsection{iLearn}
\textit{\textbf{Members:}}\\
Kun Wang$^1$ (khylon.kun.wang@gmail.com),\\
Yupeng Hu$^1$ (huyupeng@sdu.edu.cn),\\
Zhiran Li$^{1}$ (zhiranli325@gmail.com),\\
Hao Liu$^{1}$ (liuh90210@gmail.com),\\
Qianlong Xiang$^{2}$ (xiangqianlongcs@gmail.com),\\
Liqiang Nie$^{2}$ (nieliqiang@gmail.com)\\
\textit{\textbf{Affiliations:}}\\
$^1$: School of Software, Shandong University, Jinan, China\\
$^2$: School of Computer Science and Technology, Harbin Institute of Technology, Shenzhen, China

\subsection{CVSP}
\textit{\textbf{Members:}}\\
Konstantinos Chaldaiopoulos$^{1,2,3}$\\(k.chaldaiopoulos@athenarc.gr),\\
Niki Efthymiou$^{1,2,4}$ (nefthymiou@athenarc.gr),\\
Athanasia Zlatintsi$^{1,2}$ (nancy.zlatintsi@athenarc.gr),\\
Panagiotis Filntisis$^{1,2,4}$ (pfilntisis@athenarc.gr),\\
Katerina Pastra$^{3}$ (kpastra@athenarc.gr),\\
Petros Maragos$^{1,2,4}$ (petros.maragos@athenarc.gr)\\
\textit{\textbf{Affiliations:}}\\
$^1$: School of Electrical \& Computer Engineering, NTUA, Greece\\
$^2$: Robotics Institute, Athena Research Center, Greece\\
$^3$: Institute for Language and Speech Processing, Athena Research Center, Greece\\
$^4$: HERON — Hellenic Robotics Center of Excellence, Athens, Greece

\subsection{ARK\_MMLAB}
\textit{\textbf{Members:}}\\
Li Yang$^1$ (yangli.ai@bytedance.com),\\
Gen Zhan$^1$ (zhangen@bytedance.com),\\
Yiting Liao$^{1,2}$ (liaoyiting@bytedance.com),\\
Yabin Zhang$^{1,2}$ (zhangtao.ceb@bytedance.com)\\
\textit{\textbf{Affiliations:}}\\
$^1$: ByteDance Inc, China\\
$^2$: TikTok, USA

\subsection{Vertex}
\textit{\textbf{Members:}}\\
Yuxin Liu$^1$ (6243111042@stu.jiangnan.edu.cn),\\
Xu Wu$^1$ (6253111052@stu.jiangnan.edu.cn),\\
Yunheng Zheng$^{1}$ (6253115008@stu.jiangnan.edu.cn),\\
Linze Li$^{1}$(linze.li@stu.jiangnan.edu.cn),\\
Kun He$^{1}$ (6253112024@stu.jiangnan.edu.cn),\\
Cong Wu$^{1}$ (congwu@jiangnan.edu.cn),\\
Xuefeng Zhu$^{1}$ (xuefeng.zhu@jiangnan.edu.cn),\\
Tianyang Xu$^{1}$ (tianyang\_xu@163.com),\\
Xiaojun Wu$^{1}$ (wu\_xiaojun@jiangnan.edu.cn)\\
\textit{\textbf{Affiliations:}}\\
$^1$: School of Artificial Intelligence and Computer Science, Jiangnan University, China

\subsection{AAM}
\textit{\textbf{Members:}}\\
Wenzhuo Zhao$^1$ (zwz@stu.scu.edu.cn),\\
Keren Fu$^{1,2}$ (fkrsuper@scu.edu.cn)\\
\textit{\textbf{Affiliations:}}\\
$^1$: College of Computer Science, Sichuan University, China\\
$^2$: National Key Laboratory of Fundamental Science on Synthetic Vision, Sichuan University, China

\subsection{SHU-MIIPLab}
\textit{\textbf{Members:}}\\
Gongyang Li$^1$ (ligongyang@shu.edu.cn),\\
Shixiang Shi$^1$ (shishixiang@shu.edu.cn),\\
Jianlin Chen$^{1}$ (chen1026@shu.edu.cn),\\
Haibin Ling$^{2}$ (haibin.ling@gmail.com),\\
\textit{\textbf{Affiliations:}}\\
$^1$: School of Communication and Information Engineering,
Shanghai University, China\\
$^2$: Department of Artificial Intelligence, Westlake University, China

\subsection{NTR}
\textit{\textbf{Members:}}\\
Yaoxin Jiang$^{1}$ (yaoxinj2@illinois.edu),\\
Guoyi Xu$^1$ (ericx3@illinois.edu),\\
Jiajia Liu$^{1}$ (ciciliu2@illinois.edu),\\
Yaokun Shi$^{1}$ (yaokuns2@illinois.edu),\\
Jiachen Tu$^1$ (jtu9@illinois.edu)\\
\textit{\textbf{Affiliations:}}\\
$^1$: University of Illinois Urbana-Champaign, USA

\subsection{Organizers}
\textit{\textbf{Members:}}\\
Andrey Moskalenko$^{1,2,3}$ (and.v.moskalenko@gmail.com),\\
Alexey Bryncev$^{1}$ (alxbrc0@gmail.com),\\
Ivan Kosmynin$^3$ (ivan.kosmynin@graphics.cs.msu.ru),\\
Kira Shilovskaya$^3$ (kira.shilovskaya@graphics.cs.msu.ru),\\ 
Mikhail Erofeev (mikhail@erofeev.pw),\\ 
Dmitry Vatolin$^{1}$ (dmitriy@graphics.cs.msu.ru),\\
Radu Timofte$^4$ (radu.timofte@uni-wuerzburg.de)\\
\textit{\textbf{Affiliations:}}\\
$^1$: MSU Institute for Artificial Intelligence\\
$^2$: FusionBrain Lab, AXXX\\
$^3$: Lomonosov Moscow State University\\
$^4$: Computer Vision Lab, University of Würzburg

%% file: main.bib
@String(CVPR  = {CVPR})

@String(ECCV  = {ECCV})

@String(ICML  =	{ICML})

@String(ICASSP=	{ICASSP})

@String(ICIP  = {ICIP})

@String(ICLR  = {ICLR})

@String(AAAI = {AAAI})

@String(ICML = {ICML})

@String(IROS = {IEEE IROS})

@ARTICLE{shaghaghi2025focalvid,
  author={Shaghaghi, Sahand and Payne, Karissa B. and Tripp, Bryan and Dautenhahn, Kerstin and Nehaniv, Chrystopher L.},
  journal={IEEE Access}, 
  title={FocalVid: A Platform for Tracking Visual Attention to Video via Crowdsourcing Validated Against Human Gaze Data}, 
  year={2025},
  volume={13},
  number={},
  pages={159566-159581}}

@inproceedings{ntire26visage, 
  title={{ViSAGE @ NTIRE 2026 Challenge on Video Saliency Prediction: Methods and Results}}, 
  author={Wang, Kun and Hu, Yupeng and Li, Zhiran and Liu, Hao and Xiang, Qianlong and Nie, Liqiang},   
  booktitle={Proceedings of the IEEE/CVF Conference on Computer Vision and Pattern Recognition (CVPR) Workshops},  
  year = {2026} 
}

@inproceedings{chaldaiopoulos2026predjsal,
title={{PredJSal: Video Saliency via Predictive Self-Supervised Representation}},
author={Chaldaiopoulos, Konstantinos and Efthymiou, Niki and Zlatintsi, Athanasia and Filntisis, Panagiotis and Pastra, Katerina and Maragos, Petros},
booktitle={Proceedings of the IEEE/CVF Conference on Computer Vision and Pattern Recognition (CVPR) Workshops},
year = {2026}
}

@inproceedings{aim_challenge,
  title={AIM 2024 Challenge on Video Saliency Prediction: Methods and Results},
  author={Andrey Moskalenko and Alexey Bryncev and Dmitry Vatolin and Radu Timofte and Gen Zhan and Li Yang and Yunlong Tang and Yiting Liao and Jiongzhi Lin and Baitao Huang and Morteza Moradi and Mohammad Moradi and Francesco Rundo and Concetto Spampinato and Ali Borji and Simone Palazzo and others},
  booktitle={European Conference on Computer Vision},
  pages={178--194},
  year={2024},
  organization={Springer}
}

@article{zhou2023transformer,
  title={Transformer-based multi-scale feature integration network for video saliency prediction},
  author={Zhou, Xiaofei and Wu, Songhe and Shi, Ran and Zheng, Bolun and Wang, Shuai and Yin, Haibing and Zhang, Jiyong and Yan, Chenggang},
  journal={IEEE Transactions on Circuits and Systems for Video Technology},
  volume={33},
  number={12},
  pages={7696--7707},
  year={2023},
  publisher={IEEE}
}

@inproceedings{liu2022video,
  title={Video swin transformer},
  author={Liu, Ze and Ning, Jia and Cao, Yue and Wei, Yixuan and Zhang, Zheng and Lin, Stephen and Hu, Han},
  booktitle={Proceedings of the IEEE/CVF conference on computer vision and pattern recognition},
  pages={3202--3211},
  year={2022}
}

@inproceedings{xiong2024diffsal,
  title={Diffsal: Joint audio and video learning for diffusion saliency prediction},
  author={Xiong, Junwen and Zhang, Peng and You, Tao and Li, Chuanyue and Huang, Wei and Zha, Yufei},
  booktitle={Proceedings of the IEEE/CVF Conference on Computer Vision and Pattern Recognition},
  pages={27273--27283},
  year={2024}
}

@inproceedings{moradi2024salfom,
  title={Salfom: Dynamic saliency prediction with video foundation models},
  author={Moradi, Morteza and Moradi, Mohammad and Rundo, Francesco and Spampinato, Concetto and Borji, Ali and Palazzo, Simone},
  booktitle={International Conference on Pattern Recognition},
  pages={33--48},
  year={2024},
  organization={Springer}
}

@article{gu2023mamba,
  author  = {Albert Gu and Tri Dao},
  title   = {Mamba: Linear-time Sequence Modeling with Selective State Spaces},
  journal = {arXiv preprint arXiv:2312.00752},
  year    = {2023}
}

@inproceedings{ho2020ddpm,
  author    = {Jonathan Ho and Ajay Jain and Pieter Abbeel},
  title     = {Denoising Diffusion Probabilistic Models},
  booktitle = {Advances in Neural Information Processing Systems (NeurIPS)},
  pages     = {6840--6851},
  year      = {2020}
}

@article{liang2025flowanything,
  author  = {Yingping Liang and Ying Fu and Yutao Hu and Wenqi Shao and Jiaming Liu and Debing Zhang},
  title   = {Flow-Anything: Learning Real-World Optical Flow Estimation from Large-Scale Single-View Images},
  journal = {IEEE Transactions on Pattern Analysis and Machine Intelligence},
  volume  = {47},
  number  = {10},
  pages   = {8435--8452},
  year    = {2025}
}

@inproceedings{liu2024vmamba,
  author    = {Yue Liu and Yunjie Tian and Yuzhong Zhao and Hongtian Yu and Lingxi Xie and Yaowei Wang and Qixiang Ye and Jianbin Jiao and Yunfan Liu},
  title     = {VMamba: Visual State Space Model},
  booktitle = {Advances in Neural Information Processing Systems (NeurIPS)},
  pages     = {103031--103063},
  year      = {2024}
}

@inproceedings{ravi2025sam2,
  author    = {Nikhila Ravi and others},
  title     = {SAM 2: Segment Anything in Images and Videos},
  booktitle = {International Conference on Learning Representations (ICLR)},
  pages     = {28085--28128},
  year      = {2025}
}

@inproceedings{rombach2022latent,
  author    = {Robin Rombach and Andreas Blattmann and Dominik Lorenz and Patrick Esser and Bjorn Ommer},
  title     = {High-Resolution Image Synthesis with Latent Diffusion Models},
  booktitle = {Proceedings of the IEEE/CVF Conference on Computer Vision and Pattern Recognition (CVPR)},
  pages     = {10674--10685},
  year      = {2022}
}

@inproceedings{ryali2023hiera,
  author    = {Chaitanya Ryali and others},
  title     = {Hiera: A Hierarchical Vision Transformer without the Bells-and-Whistles},
  booktitle = {International Conference on Machine Learning (ICML)},
  pages     = {29441--29454},
  year      = {2023}
}

@inproceedings{tran2018r21d,
  title={A closer look at spatiotemporal convolutions for action recognition},
  author={Tran, Du and Wang, Heng and Torresani, Lorenzo and Ray, Jamie and LeCun, Yann and Paluri, Manohar},
  booktitle={Proceedings of the IEEE Conference on Computer Vision and Pattern Recognition},
  pages={6450--6459},
  year={2018}
}

@inproceedings{liu2022convnext,
  title={A ConvNet for the 2020s},
  author={Liu, Zhuang and Mao, Hanzi and Wu, Chao-Yuan and Feichtenhofer, Christoph and Darrell, Trevor and Xie, Saining},
  booktitle={Proceedings of the IEEE/CVF Conference on Computer Vision and Pattern Recognition},
  pages={11976--11986},
  year={2022}
}

@article{kay2017kinetics,
  title={The kinetics human action video dataset},
  author={Kay, Will and Carreira, Jo{\~a}o and Simonyan, Karen and Zhang, Brian and Hillier, Chloe and Vijayanarasimhan, Sudheendra and Viola, Fabio and Green, Tim and Back, Trevor and Natsev, Paul and others},
  journal={arXiv preprint arXiv:1705.06950},
  year={2017}
}

@inproceedings{deng2009imagenet,
  title={{ImageNet}: A large-scale hierarchical image database},
  author={Deng, Jia and Dong, Wei and Socher, Richard and Li, Li-Jia and Li, Kai and Fei-Fei, Li},
  booktitle={Proceedings of the IEEE Conference on Computer Vision and Pattern Recognition},
  pages={248--255},
  year={2009}
}

@inproceedings{judd2009learning,
  title={Learning to predict where humans look},
  author={Judd, Tilke and Ehinger, Krista and Durand, Fr{\'e}do and Torralba, Antonio},
  booktitle={IEEE 12th International Conference on Computer Vision},
  pages={2106--2113},
  year={2009}
}

@inproceedings{internvideo2,
  title={InternVideo2: Scaling foundation models for multimodal video understanding},
  author={Wang, Yi and Li, Kunchang and Li, Xinhao and Yu, Jiashuo and He, Yinan and Chen, Guo and Pei, Baoqi and Zheng, Rongkun and Wang, Zun and Shi, Yansong and others},
  booktitle={European conference on computer vision},
  pages={396--416},
  year={2024},
  organization={Springer}
}

@article{lora,
  title={Lora: Low-rank adaptation of large language models.},
  author={Hu, Edward J and Shen, Yelong and Wallis, Phillip and Allen-Zhu, Zeyuan and Li, Yuanzhi and Wang, Shean and Wang, Liang and Chen, Weizhu and others},
  journal={Iclr},
  volume={1},
  number={2},
  pages={3},
  year={2022}
}

@inproceedings{film,
  title={Film: Visual reasoning with a general conditioning layer},
  author={Perez, Ethan and Strub, Florian and De Vries, Harm and Dumoulin, Vincent and Courville, Aaron},
  booktitle={Proceedings of the AAAI conference on artificial intelligence},
  volume={32},
  number={1},
  year={2018}
}

@article{spratling2012,
  author  = {Spratling, M.W.},
  title   = {Predictive coding as a model of the {V1} saliency map hypothesis},
  journal = {Neural Networks},
  volume  = {26},
  pages   = {7--28},
  year    = {2012},
  doi     = {10.1016/j.neunet.2011.10.002}
}

@article{garrido2025intuitive,
  title={Intuitive physics understanding emerges from self-supervised pretraining on natural videos},
  author={Garrido, Quentin and Ballas, Nicolas and Assran, Mahmoud and Bardes, Adrien and Najman, Laurent and Rabbat, Michael and Dupoux, Emmanuel and LeCun, Yann},
  journal={arXiv preprint arXiv:2502.11831},
  year={2025}
}

@article{assran2025v,
  title={V-jepa 2: Self-supervised video models enable understanding, prediction and planning},
  author={Assran, Mido and Bardes, Adrien and Fan, David and Garrido, Quentin and Howes, Russell and Muckley, Matthew and Rizvi, Ammar and Roberts, Claire and Sinha, Koustuv and Zholus, Artem and others},
  journal={arXiv preprint arXiv:2506.09985},
  year={2025}
}

@article{rao1999predictive,
  title={Predictive coding in the visual cortex: a functional interpretation of some extra-classical receptive-field effects},
  author={Rao, Rajesh PN and Ballard, Dana H},
  journal={Nature neuroscience},
  volume={2},
  number={1},
  pages={79--87},
  year={1999},
  publisher={Nature Publishing Group}
}

@inproceedings{wang2024internvideo2,
  title={Internvideo2: Scaling foundation models for multimodal video understanding},
  author={Wang, Yi and Li, Kunchang and Li, Xinhao and Yu, Jiashuo and He, Yinan and Chen, Guo and Pei, Baoqi and Zheng, Rongkun and Wang, Zun and Shi, Yansong and others},
  booktitle={European conference on computer vision},
  pages={396--416},
  year={2024},
  organization={Springer}
}

@article{simeoni2025dinov3,
  title={Dinov3},
  author={Sim{\'e}oni, Oriane and Vo, Huy V and Seitzer, Maximilian and Baldassarre, Federico and Oquab, Maxime and Jose, Cijo and Khalidov, Vasil and Szafraniec, Marc and Yi, Seungeun and Ramamonjisoa, Micha{\"e}l and others},
  journal={arXiv preprint arXiv:2508.10104},
  year={2025}
}

@inproceedings{radford2021learning,
  title={Learning transferable visual models from natural language supervision},
  author={Radford, Alec and Kim, Jong Wook and Hallacy, Chris and Ramesh, Aditya and Goh, Gabriel and Agarwal, Sandhini and Sastry, Girish and Askell, Amanda and Mishkin, Pamela and Clark, Jack and others},
  booktitle=ICML,
  pages={8748--8763},
  year={2021},
  organization={PmLR}
}

@inproceedings{wu2022wav2clip,
    title={Wav2CLIP: Learning Robust Audio Representations From CLIP},
    author={Wu, Ho-Hsiang and Seetharaman, Prem and Kumar, Kundan and Bello, Juan Pablo},
    booktitle=ICASSP,
    year={2022}
}

@INPROCEEDINGS {gitman_semiautomatic,
    AUTHOR    = "Yury Gitman and Mikhail Erofeev and Dmitriy Vatolin
                 and Andrey Bolshakov and Alexey Fedorov",
    TITLE     = "Semiautomatic Visual-Attention Modeling and Its 
                 Application to Video Compression",
    BOOKTITLE = "2014 IEEE International Conference on Image Processing
                 (ICIP) (ICIP 2014)",
    ADDRESS   = "Paris, France",
    PAGES     = "1105-1109",
    DAYS      =  27,
    MONTH     =  oct,
    YEAR      =  2014,
  }

@inproceedings{lyudvichenko2017semiautomatic,
  title={A semiautomatic saliency model and its application to video compression},
  author={Lyudvichenko, Vitaliy and Erofeev, Mikhail and Gitman, Yury and Vatolin, Dmitriy},
  booktitle={2017 13th IEEE International Conference on Intelligent Computer Communication and Processing (ICCP)},
  pages={403--410},
  year={2017},
  organization={IEEE}
}

@inproceedings{patel2021saliency,
  title={Saliency driven perceptual image compression},
  author={Patel, Yash and Appalaraju, Srikar and Manmatha, R},
  booktitle={Proceedings of the IEEE/CVF winter conference on applications of computer vision},
  pages={227--236},
  year={2021}
}

@inproceedings{alexey2025bridging,
  title={Bridging the gap between saliency prediction and image quality assessment},
  author={Alexey, Kirillov and Moskalenko, Andrey and Vatolin, Dmitriy},
  booktitle={2025 33rd European Signal Processing Conference (EUSIPCO)},
  pages={656--660},
  year={2025},
  organization={IEEE}
}

@inproceedings{yang2019sgdnet,
  title={SGDNet: An end-to-end saliency-guided deep neural network for no-reference image quality assessment},
  author={Yang, Sheng and Jiang, Qiuping and Lin, Weisi and Wang, Yongtao},
  booktitle={Proceedings of the 27th ACM international conference on multimedia},
  pages={1383--1391},
  year={2019}
}

@article{hadizadeh2013saliency,
  title={Saliency-aware video compression},
  author={Hadizadeh, Hadi and Baji{\'c}, Ivan V},
  journal={IEEE Transactions on Image Processing},
  volume={23},
  number={1},
  pages={19--33},
  year={2013},
  publisher={IEEE}
}

@article{chang2026saliency,
  title={Saliency-guided video coding via recurrent learning and perceptual quality assessment},
  author={Chang, Tz-Cheng and Hsiao, Hsu-Feng},
  journal={Signal Processing: Image Communication},
  pages={117536},
  year={2026},
  publisher={Elsevier}
}

@inproceedings{qu2025kvq,
  title={KVQ: boosting video quality assessment via saliency-guided local perception},
  author={Qu, Yunpeng and Yuan, Kun and Xie, Qizhi and Sun, Ming and Zhou, Chao and Wang, Jian},
  booktitle={Proceedings of the Computer Vision and Pattern Recognition Conference},
  pages={2150--2160},
  year={2025}
}

@article{zhang2015application,
  title={The application of visual saliency models in objective image quality assessment: A statistical evaluation},
  author={Zhang, Wei and Borji, Ali and Wang, Zhou and Le Callet, Patrick and Liu, Hantao},
  journal={IEEE transactions on neural networks and learning systems},
  volume={27},
  number={6},
  pages={1266--1278},
  year={2015},
  publisher={IEEE}
}

@article{fang2012saliency,
  title={Saliency detection in the compressed domain for adaptive image retargeting},
  author={Fang, Yuming and Chen, Zhenzhong and Lin, Weisi and Lin, Chia-Wen},
  journal={IEEE Transactions on Image Processing},
  volume={21},
  number={9},
  pages={3888--3901},
  year={2012},
  publisher={IEEE}
}

@inproceedings{miangoleh2023realistic,
  title={Realistic saliency guided image enhancement},
  author={Miangoleh, S Mahdi H and Bylinskii, Zoya and Kee, Eric and Shechtman, Eli and Aksoy, Ya{\u{g}}iz},
  booktitle={Proceedings of the IEEE/CVF Conference on Computer Vision and Pattern Recognition},
  pages={186--194},
  year={2023}
}

@article{ahmadi2021context,
  title={Context-aware saliency detection for image retargeting using convolutional neural networks},
  author={Ahmadi, Mahdi and Karimi, Nader and Samavi, Shadrokh},
  journal={Multimedia Tools and Applications},
  volume={80},
  number={8},
  pages={11917--11941},
  year={2021},
  publisher={Springer}
}

@article{guo2024irnet,
  title={IRNet-RS: image retargeting network via relative saliency},
  author={Guo, Yingchun and Zhang, Meng and Hao, Xiaoke and Yan, Gang},
  journal={Neural Computing and Applications},
  volume={36},
  number={8},
  pages={4133--4149},
  year={2024},
  publisher={Springer}
}

@inproceedings{yun2022panoramic,
  title={Panoramic vision transformer for saliency detection in 360 videos},
  author={Yun, Heeseung and Lee, Sehun and Kim, Gunhee},
  booktitle={European Conference on Computer Vision},
  pages={422--439},
  year={2022},
  organization={Springer}
}

@article{cokelek2025spherical,
  title={Spherical Vision Transformers for Audio-Visual Saliency Prediction in 360 Videos},
  author={Cokelek, Mert and Ozsoy, Halit and Imamoglu, Nevrez and Ozcinar, Cagri and Ayhan, Inci and Erdem, Erkut and Erdem, Aykut},
  journal={IEEE transactions on pattern analysis and machine intelligence},
  year={2025},
  publisher={IEEE}
}

@inproceedings{wahba2025enhancement,
  title={Enhancement of 360° Video Streaming through Saliency-Guided Viewport Prediction},
  author={Wahba, Mahmoud ZA},
  booktitle={2025 International Conference on Visual Communications and Image Processing (VCIP)},
  pages={1--3},
  year={2025},
  organization={IEEE}
}

@article{jiao2026diffgaze,
  title={Diffgaze: A diffusion model for modelling fine-grained human gaze behaviour on 360 images},
  author={Jiao, Chuhan and Wang, Yao and Zhang, Guanhua and B{\^a}ce, Mihai and Hu, Zhiming and Bulling, Andreas},
  journal={ACM Transactions on Interactive Intelligent Systems},
  volume={16},
  number={1},
  pages={1--23},
  year={2026},
  publisher={ACM New York, NY}
}

@inproceedings{wang2022salientvr,
  title={SalientVR: Saliency-driven mobile 360-degree video streaming with gaze information},
  author={Wang, Shibo and Yang, Shusen and Li, Hailiang and Zhang, Xiaodan and Zhou, Chen and Xu, Chenren and Qian, Feng and Wang, Nanbin and Xu, Zongben},
  booktitle={Proceedings of the 28th Annual International Conference on Mobile Computing And Networking},
  pages={542--555},
  year={2022}
}

@article{zhao2026rate,
  title={Rate Control for 360° Versatile Video Coding Based on Visual Gaze Mechanism},
  author={Zhao, Zeming and Wang, Meng and Sui, Xiangjie and Chen, Peilin and He, Xiaohai and Wang, Shiqi},
  journal={IEEE Transactions on Multimedia},
  year={2026},
  publisher={IEEE}
}

@inproceedings{zhang2018saliency,
  title={Saliency detection in 360 videos},
  author={Zhang, Ziheng and Xu, Yanyu and Yu, Jingyi and Gao, Shenghua},
  booktitle={Proceedings of the European conference on computer vision (ECCV)},
  pages={488--503},
  year={2018}
}

@article{lee2005mesh,
author = {Lee, Chang Ha and Varshney, Amitabh and Jacobs, David W.},
title = {Mesh saliency},
year = {2005},
issue_date = {July 2005},
publisher = {Association for Computing Machinery},
address = {New York, NY, USA},
volume = {24},
number = {3},
issn = {0730-0301},
url = {https://doi.org/10.1145/1073204.1073244},
doi = {10.1145/1073204.1073244},
journal = {ACM Trans. Graph.},
month = jul,
pages = {659–666},
numpages = {8}
}

@article{ding2023towards,
  title={Towards 3d colored mesh saliency: Database and benchmarks},
  author={Ding, Xiaoying and Chen, Zhao and Lin, Weisi and Chen, Zhenzhong},
  journal={IEEE Transactions on Multimedia},
  volume={26},
  pages={3580--3591},
  year={2023},
  publisher={IEEE}
}

@inproceedings{zhang2025textured,
  title={Textured mesh saliency: Bridging geometry and texture for human perception in 3d graphics},
  author={Zhang, Kaiwei and Zhu, Dandan and Min, Xiongkuo and Zhai, Guangtao},
  booktitle={Proceedings of the AAAI Conference on Artificial Intelligence},
  volume={39},
  number={9},
  pages={9977--9984},
  year={2025}
}

@article{martin2024sal3d,
  title={Sal3d: a model for saliency prediction in 3d meshes},
  author={Martin, Daniel and Fandos, Andres and Masia, Belen and Serrano, Ana},
  journal={The Visual Computer},
  volume={40},
  number={11},
  pages={7761--7771},
  year={2024},
  publisher={Springer}
}

@inproceedings{zhang2025mesh,
  title={Mesh Mamba: A unified state space model for saliency prediction in non-textured and textured meshes},
  author={Zhang, Kaiwei and Zhu, Dandan and Min, Xiongkuo and Zhai, Guangtao},
  booktitle={Proceedings of the Computer Vision and Pattern Recognition Conference},
  pages={16219--16228},
  year={2025}
}

@article{nousias2023deep,
  title={Deep saliency mapping for 3D meshes and applications},
  author={Nousias, Stavros and Arvanitis, Gerasimos and Lalos, Aris and Moustakas, Konstantinos},
  journal={ACM Transactions on Multimedia Computing, Communications and Applications},
  volume={19},
  number={2},
  pages={1--22},
  year={2023},
  publisher={ACM New York, NY}
}

@article{dos2023saliency,
title = {Saliency detection for large-scale mesh decimation},
journal = {Computers \& Graphics},
volume = {111},
pages = {63-76},
year = {2023},
issn = {0097-8493},
doi = {https://doi.org/10.1016/j.cag.2023.01.012},
url = {https://www.sciencedirect.com/science/article/pii/S0097849323000134},
author = {Rafael {Kuffner dos Anjos} and Richard Andrew Roberts and Benjamin Allen and Joaquim Jorge and Ken Anjyo},
}

@article{itti1998model,
  title={A model of saliency-based visual attention for rapid scene analysis},
  author={Itti, Laurent and Koch, Christof and Niebur, Ernst},
  journal={IEEE Transactions on pattern analysis and machine intelligence},
  volume={20},
  number={11},
  pages={1254--1259},
  year={1998},
  publisher={Ieee}
}

@article{harel2006graph,
  title={Graph-based visual saliency},
  author={Harel, Jonathan and Koch, Christof and Perona, Pietro},
  journal={Advances in neural information processing systems},
  volume={19},
  year={2006}
}

@inproceedings{guo2008spatio,
  title={Spatio-temporal saliency detection using phase spectrum of quaternion fourier transform},
  author={Guo, Chenlei and Ma, Qi and Zhang, Liming},
  booktitle={2008 IEEE conference on computer vision and pattern recognition},
  pages={1--8},
  year={2008},
  organization={IEEE}
}

@article{mahadevan2009spatiotemporal,
  title={Spatiotemporal saliency in dynamic scenes},
  author={Mahadevan, Vijay and Vasconcelos, Nuno},
  journal={IEEE transactions on pattern analysis and machine intelligence},
  volume={32},
  number={1},
  pages={171--177},
  year={2009},
  publisher={IEEE}
}

@article{marat2009modelling,
  title={Modelling spatio-temporal saliency to predict gaze direction for short videos},
  author={Marat, Sophie and Ho Phuoc, Tien and Granjon, Lionel and Guyader, Nathalie and Pellerin, Denis and Gu{\'e}rin-Dugu{\'e}, Anne},
  journal={International journal of computer vision},
  volume={82},
  number={3},
  pages={231--243},
  year={2009},
  publisher={Springer}
}

@article{kroner2020contextual,
  title={Contextual encoder--decoder network for visual saliency prediction},
  author={Kroner, Alexander and Senden, Mario and Driessens, Kurt and Goebel, Rainer},
  journal={Neural Networks},
  volume={129},
  pages={261--270},
  year={2020},
  publisher={Elsevier}
}

@article{lou2022455,
title = {TranSalNet: Towards perceptually relevant visual saliency prediction},
journal = {Neurocomputing},
volume = {494},
pages = {455-467},
year = {2022},
issn = {0925-2312},
doi = {https://doi.org/10.1016/j.neucom.2022.04.080},
url = {https://www.sciencedirect.com/science/article/pii/S0925231222004714},
author = {Jianxun Lou and Hanhe Lin and David Marshall and Dietmar Saupe and Hantao Liu},
}

@inproceedings{xiong2023casp,
  title={CASP-Net: Rethinking video saliency prediction from an audio-visual consistency perceptual perspective},
  author={Xiong, Junwen and Wang, Ganglai and Zhang, Peng and Huang, Wei and Zha, Yufei and Zhai, Guangtao},
  booktitle={Proceedings of the IEEE/CVF conference on computer vision and pattern recognition},
  pages={6441--6450},
  year={2023}
}

@inproceedings{drostejiao2020,
     author = {{Droste}, Richard and {Jiao}, Jianbo and {Noble}, J. Alison},
      title = "{Unified Image and Video Saliency Modeling}",
  booktitle = {Proceedings of the 16th European Conference on Computer Vision (ECCV)},
       year = {2020},
}

@inproceedings{jain2021vinet,
  title={Vinet: Pushing the limits of visual modality for audio-visual saliency prediction},
  author={Jain, Samyak and Yarlagadda, Pradeep and Jyoti, Shreyank and Karthik, Shyamgopal and Subramanian, Ramanathan and Gandhi, Vineet},
  booktitle={2021 IEEE/RSJ International Conference on Intelligent Robots and Systems (IROS)},
  pages={3520--3527},
  year={2021},
  organization={IEEE}
}

@article{tavakoli2019dave,
  title={Dave: A deep audio-visual embedding for dynamic saliency prediction},
  author={Tavakoli, Hamed R and Borji, Ali and Rahtu, Esa and Kannala, Juho},
  journal={arXiv preprint arXiv:1905.10693},
  year={2019}
}

@article{chen2022comprehensive,
  title={A comprehensive survey on video saliency detection with auditory information: The audio-visual consistency perceptual is the key!},
  author={Chen, Chenglizhao and Song, Mengke and Song, Wenfeng and Guo, Li and Jian, Muwei},
  journal={IEEE Transactions on Circuits and Systems for Video Technology},
  volume={33},
  number={2},
  pages={457--477},
  year={2022},
  publisher={IEEE}
}

@inproceedings{tang2025cardiff,
  title={Cardiff: Video salient object ranking chain of thought reasoning for saliency prediction with diffusion},
  author={Tang, Yunlong and Zhan, Gen and Yang, Li and Liao, Yiting and Xu, Chenliang},
  booktitle={Proceedings of the AAAI Conference on Artificial Intelligence},
  volume={39},
  number={7},
  pages={7302--7310},
  year={2025}
}

@inproceedings{chen2025explainable,
  title={Explainable Saliency: Articulating Reasoning with Contextual Prioritization},
  author={Chen, Nuo and Jiang, Ming and Zhao, Qi},
  booktitle={Proceedings of the Computer Vision and Pattern Recognition Conference},
  pages={9601--9610},
  year={2025}
}

@article{mathe2014actions,
  title={Actions in the eye: Dynamic gaze datasets and learnt saliency models for visual recognition},
  author={Mathe, Stefan and Sminchisescu, Cristian},
  journal={IEEE transactions on pattern analysis and machine intelligence},
  volume={37},
  number={7},
  pages={1408--1424},
  year={2014},
  publisher={IEEE}
}

@inproceedings{wang2018revisiting,
    title={Revisiting Video Saliency: A Large-scale Benchmark and a New Model},
    author={Wang, Wenguan and Shen, Jianbing and Guo, Fang and Cheng, Ming-Ming and Borji, Ali},
    booktitle={Proceedings of the IEEE Conference on Computer Vision and Pattern Recognition},
    year={2018},
}

@inproceedings{jiang2015salicon,
  title={Salicon: Saliency in context},
  author={Jiang, Ming and Huang, Shengsheng and Duan, Juanyong and Zhao, Qi},
  booktitle={Proceedings of the IEEE conference on computer vision and pattern recognition},
  pages={1072--1080},
  year={2015}
}

@article{kim2017bubbleview,
  title={Bubbleview: an interface for crowdsourcing image importance maps and tracking visual attention},
  author={Kim, Nam Wook and Bylinskii, Zoya and Borkin, Michelle A and Gajos, Krzysztof Z and Oliva, Aude and Durand, Fredo and Pfister, Hanspeter},
  journal={ACM Transactions on Computer-Human Interaction (TOCHI)},
  volume={24},
  number={5},
  pages={1--40},
  year={2017},
  publisher={ACM New York, NY, USA}
}

@inproceedings{tavakoli2017saliency,
  title={Saliency revisited: Analysis of mouse movements versus fixations},
  author={Tavakoli, Hamed R and Ahmed, Fawad and Borji, Ali and Laaksonen, Jorma},
  booktitle={Proceedings of the ieee conference on computer vision and pattern recognition},
  pages={1774--1782},
  year={2017}
}

@INPROCEEDINGS{lyudvichenko2019predicting,
  title={Predicting video saliency using crowdsourced mouse-tracking data},
  author={Lyudvichenko, Vitaliy and Vatolin, Dmitriy},
  BOOKTITLE={Proceedings of the 29th International Conference on Computer Graphics and Vision},
  year={2019},
  PAGES     = {127--130},
  volume={2485},
  doi={http://dx.doi.org/10.30987/graphicon-2019-2-127-130}
}

@inproceedings{wang2024youtube,
  title={YouTube SFV+ HDR quality dataset},
  author={Wang, Yilin and Yim, Joong Gon and Birkbeck, Neil and Adsumilli, Balu},
  booktitle={2024 IEEE International Conference on Image Processing (ICIP)},
  pages={96--102},
  year={2024},
  organization={IEEE}
}

@misc{Farre2024FineVideo,
  title={FineVideo},
  author={Farré, Miquel and Marafioti, Andi and Tunstall, Lewis and Von Werra, Leandro and Wolf, Thomas},
  year={2024},
  howpublished={\url{https://huggingface.co/datasets/HuggingFaceFV/finevideo}},
}

@article{bylinskii2018different,
  title={What do different evaluation metrics tell us about saliency models?},
  author={Bylinskii, Zoya and Judd, Tilke and Oliva, Aude and Torralba, Antonio and Durand, Fr{\'e}do},
  journal={IEEE transactions on pattern analysis and machine intelligence},
  volume={41},
  number={3},
  pages={740--757},
  year={2018},
  publisher={IEEE}
}

@inproceedings{safonov2025ntire,
  title={NTIRE 2025 challenge on UGC video enhancement: Methods and results},
  author={Safonov, Nickolay and Bryntsev, Alexey and Moskalenko, Andrey and Kulikov, Dmitry and Vatolin, Dmitriy and Timofte, Radu and Lei, Haibo and Gao, Qifan and Luo, Qing and Li, Yaqing and others},
  booktitle={Proceedings of the Computer Vision and Pattern Recognition Conference},
  pages={1503--1513},
  year={2025}
}

@inproceedings{ntire26deepfake, 
title={{    Robust Deepfake Detection, NTIRE 2026 Challenge: Report    }}, 
author={    Hopf, Benedikt and  Timofte, Radu and others    },   
booktitle={Proceedings of the IEEE/CVF Conference on Computer Vision and Pattern Recognition (CVPR) Workshops},  
year = {2026} 
}

@inproceedings{ntire26hrdepth, 
title={{    NTIRE 2026 Challenge on High-Resolution Depth of non-Lambertian Surfaces    }}, 
author={    Zama Ramirez, Pierluigi and  Tosi, Fabio and  Di Stefano, Luigi and  Timofte, Radu and  Costanzino, Alex and  Poggi, Matteo and  Salti, Samuele and  Mattoccia, Stefano and others    },   
booktitle={Proceedings of the IEEE/CVF Conference on Computer Vision and Pattern Recognition (CVPR) Workshops},  
year = {2026} 
}

@inproceedings{ntire26raim_fusion, 
title={{    NTIRE 2026 The 3rd Restore Any Image Model (RAIM) Challenge: Multi-Exposure Image Fusion in Dynamic Scenes (Track2)    }}, 
author={    Qu, Lishen and  Liu, Yao and  Liang, Jie and  Zeng, Hui and  Dai, Wen and  Guan, Ya-nan and  Qin, Guanyi and  Zhou, Shihao and  Yang, Jufeng and  Zhang, Lei and  Timofte, Radu and others    },   
booktitle={Proceedings of the IEEE/CVF Conference on Computer Vision and Pattern Recognition (CVPR) Workshops},  
year = {2026} 
}

@inproceedings{ntire26raim_portrait, 
title={{    NTIRE 2026 The 3rd Restore Any Image Model (RAIM) Challenge: AI Flash Portrait (Track 3)    }}, 
author={    Guan, Ya-nan and  Zhang, Shaonan and  Guo, Hang and  Wang, Yawen and  Fan, Xinying and  Liang, Jie and  Zeng, Hui and  Qin, Guanyi and  Qu, Lishen and  Dai, Tao and  Xia, Shu-Tao and  Zhang, Lei and  Timofte, Radu and others    },   
booktitle={Proceedings of the IEEE/CVF Conference on Computer Vision and Pattern Recognition (CVPR) Workshops},  
year = {2026} 
}

@inproceedings{ntire26raim_piqa, 
title={{    NTIRE 2026 The 3rd Restore Any Image Model (RAIM) Challenge: Professional Image Quality Assessment (Track 1)    }}, 
author={    Qin, Guanyi and  Liang, Jie and  Zhang, Bingbing and  Qu, Lishen and  Guan, Ya-nan and  Zeng, Hui and  Zhang, Lei and  Timofte, Radu and others    },   
booktitle={Proceedings of the IEEE/CVF Conference on Computer Vision and Pattern Recognition (CVPR) Workshops},  
year = {2026} 
}

@inproceedings{ntire26lightsr, 
title={{    NTIRE 2026 Challenge on Light Field Image Super-Resolution: Methods and Results    }}, 
author={    Wang, Yingqian and  Liang, Zhengyu and  Zhang, Fengyuan and  Zhao, Wending and  Wang, Longguang and  Li, Juncheng and  Yang, Jungang and  Timofte, Radu and  Guo, Yulan and others    },   
booktitle={Proceedings of the IEEE/CVF Conference on Computer Vision and Pattern Recognition (CVPR) Workshops},  
year = {2026} 
}

@inproceedings{ntire263dsr, 
title={{    NTIRE 2026 Challenge on 3D Content Super-Resolution: Methods and Results    }}, 
author={    Wang, Longguang and  Guo, Yulan and  Wang, Yingqian and  Li, Juncheng and  Peng, Sida and  Zhang, Ye and  Timofte, Radu and  Chen, Minglin and  Wang, Yi and  Hu, Qibin and  Lei, Wenjie and others    },   
booktitle={Proceedings of the IEEE/CVF Conference on Computer Vision and Pattern Recognition (CVPR) Workshops},  
year = {2026} 
}

@inproceedings{ntire26videores, 
title={{    NTIRE 2026 Challenge on Bitstream-Corrupted Video Restoration: Methods and Results    }}, 
author={    Zou, Wenbin and  Liu, Tianyi and  Wu, Kejun and  Zhuang, Huiping and  Wu, Zongwei and  Zhou, Zhuyun and  Timofte, Radu and  others     },   booktitle={Proceedings of the IEEE/CVF Conference on Computer Vision and Pattern Recognition (CVPR) Workshops},  
year = {2026} 
}

@inproceedings{ntire26XAIGCqa, 
title={{    NTIRE 2026 X-AIGC Quality Assessment Challenge: Methods and Results    }}, 
author={    Liu, Xiaohong and  Min, Xiongkuo and  Zhai, Guangtao and  Hu, Qiang and  Cao, Jiezhang and  Zhou, Yu and  Sun, Wei and  Wen, Farong and  Xu, Zitong and  Zhou, Yingjie and  Duan, Huiyu and  Liu, Lu and  Wang, Jiarui and  Luo, Siqi and  Li, Chunyi and  Xu, Li and  Zhang, Zicheng and  Shi, Yue and  Wang, Yubo and  Zhang, Minghong and  Guo, Chunchao and  Hu, Zhichao and  Chen, Mingtao and  Wu, Xiele and  Ma, Xin and  Lv, Zhaohe and  Xue, Yuanhao and  Wang, Jiaqi and  Sha, Xinxing and  Timofte, Radu and  others    },   
booktitle={Proceedings of the IEEE/CVF Conference on Computer Vision and Pattern Recognition (CVPR) Workshops},  
year = {2026} 
}

@inproceedings{ntire26shadow, 
title={{    Advances in Single-Image Shadow Removal: Results from the NTIRE 2026 Challenge    }}, 
author={    Vasluianu, Florin-Alexandru and  Seizinger, Tim and  Zhou, Zhuyun and  Wu, Zongwei and  Timofte, Radu and  others     },   
booktitle={Proceedings of the IEEE/CVF Conference on Computer Vision and Pattern Recognition (CVPR) Workshops},  
year = {2026} 
}

@inproceedings{ntire26lightnorm, 
title={{    Learning-Based Ambient Lighting Normalization: NTIRE 2026 Challenge Results and Findings    }}, 
author={    Vasluianu, Florin-Alexandru and  Seizinger, Tim and  Chen, Jeffrey and  Zhou, Zhuyun and  Wu, Zongwei and  Timofte, Radu and  others    },   booktitle={Proceedings of the IEEE/CVF Conference on Computer Vision and Pattern Recognition (CVPR) Workshops},  
year = {2026} 
}

@inproceedings{ntire26bokeh, 
title={{    The First Controllable Bokeh Rendering Challenge at NTIRE 2026    }}, 
author={    Seizinger, Tim and  Vasluianu, Florin-Alexandru and  Conde, Marcos V. and  Chen, Jeffrey and  Zhou, Zhuyun and  Wu, Zongwei and  Timofte, Radu and  others    },   
booktitle={Proceedings of the IEEE/CVF Conference on Computer Vision and Pattern Recognition (CVPR) Workshops},  
year = {2026} 
}

@inproceedings{ntire26ripdetseg, 
title={{    NTIRE 2026 Rip Current Detection and Segmentation (RipDetSeg) Challenge Report    }}, 
author={    Dumitriu, Andrei and  Ralhan, Aakash and  Miron, Florin and  Tatui, Florin and  Ionescu, Radu Tudor and  Timofte, Radu and  others     },   booktitle={Proceedings of the IEEE/CVF Conference on Computer Vision and Pattern Recognition (CVPR) Workshops},  
year = {2026} 
}

@inproceedings{ntire26llie, 
title={{    Low Light Image Enhancement Challenge at NTIRE 2026    }}, 
author={    Ciubotariu, George and  S M A,  Sharif and  Rehman, Abdur and  Ali Dharejo, Fayaz and  Naqvi, Rizwan Ali and  Conde, Marcos and  Timofte, Radu and others    },   
booktitle={Proceedings of the IEEE/CVF Conference on Computer Vision and Pattern Recognition (CVPR) Workshops},  
year = {2026} 
}

@inproceedings{ntire26highfps, 
title={{    High FPS Video Frame Interpolation Challenge at NTIRE 2026    }}, 
author={    Ciubotariu, George and  Zhou, Zhuyun and  Jin, Yeying and  Wu, Zongwei and  Timofte, Radu and  others    },   
booktitle={Proceedings of the IEEE/CVF Conference on Computer Vision and Pattern Recognition (CVPR) Workshops},  
year = {2026} 
}

@inproceedings{ntire26nthaze, 
title={{    NT-HAZE: A Benchmark Dataset for Realistic Night-time Image Dehazing    }}, 
author={    Ancuti, Radu and  Ancuti, Codruta and  Timofte, Radu and  Ancuti, Cosmin    },   
booktitle={Proceedings of the IEEE/CVF Conference on Computer Vision and Pattern Recognition (CVPR) Workshops},  
year = {2026} 
}

@inproceedings{ntire26nthaze_rep, 
title={{    NTIRE 2026 Nighttime Image Dehazing Challenge Report    }}, 
author={    Ancuti, Radu and  Brateanu, Alexandru and  Vasluianu, Florin and  Balmez, Raul and  Orhei, Ciprian and  Ancuti, Codruta and  Timofte, Radu and  Ancuti, Cosmin and others    },   
booktitle={Proceedings of the IEEE/CVF Conference on Computer Vision and Pattern Recognition (CVPR) Workshops},  
year = {2026} 
}

@inproceedings{ntire26isp, 
title={{    NTIRE 2026 Challenge on Learned Smartphone ISP with Unpaired Data: Methods and Results    }}, 
author={    Perevozchikov, Georgy and  Vladimirov, Daniil and  Timofte, Radu and  others    },   
booktitle={Proceedings of the IEEE/CVF Conference on Computer Vision and Pattern Recognition (CVPR) Workshops},  
year = {2026} 
}

@inproceedings{ntire26ugcvideo, 
title={{    NTIRE 2026 Challenge on Short-form UGC Video Restoration in the Wild with Generative Models: Datasets, Methods and Results    }}, author={    Li, Xin and  Gong, Jiachao and  Wang, Xijun and  Xiong, Shiyao and  Li, Bingchen and  Yao, Suhang  and  Zhou, Chao and  Chen, Zhibo and  Timofte, Radu and others    },   
booktitle={Proceedings of the IEEE/CVF Conference on Computer Vision and Pattern Recognition (CVPR) Workshops},  
year = {2026} 
}

@inproceedings{ntire26dual_focus, 
title={{    NTIRE 2026 The Second Challenge on Day and Night Raindrop Removal for Dual-Focused Images: Methods and Results    }}, 
author={    Li, Xin and  Jin, Yeying and  Yao, Suhang and  Lin, Beibei and  Fan, Zhaoxin and   Yan, Wending and  Jin, Xin and  Wu, Zongwei  and  Li, Bingchen  and  Shi, Peishu and  Yang, Yufei and  Li, Yu and  Chen, Zhibo  and  Wen, Bihan and  Tan, Robby and  Timofte, Radu and others    },   
booktitle={Proceedings of the IEEE/CVF Conference on Computer Vision and Pattern Recognition (CVPR) Workshops},  
year = {2026} 
}

@inproceedings{ntire26srx4, 
title={{    The Fourth Challenge on Image Super-Resolution (×4) at NTIRE 2026: Benchmark Results and Method Overview    }}, 
author={    Chen, Zheng and  Liu, Kai and  Wang, Jingkai and  Yan, Xianglong and  Li, Jianze and  Zhang, Ziqing and  Gong, Jue and  Li, Jiatong and  Sun, Lei and  Liu, Xiaoyang and  Timofte, Radu and  Zhang, Yulun and others    },   
booktitle={Proceedings of the IEEE/CVF Conference on Computer Vision and Pattern Recognition (CVPR) Workshops},  
year = {2026} 
}

@inproceedings{ntire26retouching, 
title={{    Photography Retouching Transfer, NTIRE 2026 Challenge: Report    }}, 
author={    Elezabi, Omar and  V. Conde, Marcos and  Wu, Zongwei and  Jin, Yeying and  Timofte, Radu and others    },   
booktitle={Proceedings of the IEEE/CVF Conference on Computer Vision and Pattern Recognition (CVPR) Workshops},  
year = {2026} 
}

@inproceedings{ntire26rwsr, 
title={{    The First Challenge on Mobile Real-World Image Super-Resolution at NTIRE 2026: Benchmark Results and Method Overview    }}, 
author={    Li, Jiatong and  Chen, Zheng and  Liu, Kai and  Wang, Jingkai and  Zhou, Zihan and  Liu, Xiaoyang and  Zhu, Libo and  Timofte, Radu and  Zhang, Yulun and others    },   
booktitle={Proceedings of the IEEE/CVF Conference on Computer Vision and Pattern Recognition (CVPR) Workshops},  
year = {2026} 
}

@inproceedings{ntire26rsirsr, 
title={{    The First Challenge on Remote Sensing Infrared Image Super-Resolution at NTIRE 2026: Benchmark Results and Method Overview    }}, author={    Liu, Kai and  Yue, Haoyang and  Lin, Zeli and  Chen, Zheng and  Wang, Jingkai and  Gong, Jue and  Timofte, Radu and  Zhang, Yulun and  others    },   
booktitle={Proceedings of the IEEE/CVF Conference on Computer Vision and Pattern Recognition (CVPR) Workshops},  
year = {2026} 
}

@inproceedings{ntire26aigendet, 
title={{    NTIRE 2026 Challenge on Robust AI-Generated Image Detection in the Wild    }}, 
author={    Gushchin, Aleksandr and  Abud, Khaled and  Shumitskaya, Ekaterina and  Filippov, Artem and  Bychkov, Georgii and  Lavrushkin, Sergey and  Erofeev, Mikhail and  Antsiferova, Anastasia and  Chen, Changsheng and  Tan, Shunquan and  Timofte, Radu and  Vatolin, Dmitriy and others    },
booktitle={Proceedings of the IEEE/CVF Conference on Computer Vision and Pattern Recognition (CVPR) Workshops},  
year = {2026} 
}

@inproceedings{ntire26cdfsod, 
title={{    The Second Challenge on Cross-Domain Few-Shot Object Detection at NTIRE 2026: Methods and Results    }}, 
author={    Qiu, Xingyu and  Fu, Yuqian and  Geng, Jiawei and  Ren, Bin and  Pan, Jiancheng and  Wu, Zongwei and  Tang, Hao and  Fu, Yanwei and  Timofte, Radu and  Sebe, Nicu and  Elhoseiny, Mohamed and others    },   
booktitle={Proceedings of the IEEE/CVF Conference on Computer Vision and Pattern Recognition (CVPR) Workshops},  
year = {2026} 
}

@inproceedings{ntire26finrec, 
title={{    NTIRE 2026 Challenge on End-to-End Financial Receipt Restoration and Reasoning from Degraded Images: Datasets, Methods and Results    }}, author={    Guan, Bochen and  Li, Jinlong and  Yang, Kangning and  Ke, Chuang and  Cai, Jie and  Vasluianu, Florin and  Timofte, Radu and others    },   booktitle={Proceedings of the IEEE/CVF Conference on Computer Vision and Pattern Recognition (CVPR) Workshops},  
year = {2026} 
}

@inproceedings{ntire26faceres, 
title={{    The Second Challenge on Real-World Face Restoration at NTIRE 2026: Methods and Results    }}, 
author={    Wang, Jingkai and  Gong, Jue and  Chen, Zheng and  Liu, Kai and  Li, Jiatong and  Zhang, Yulun and  Timofte, Radu and  others    },
booktitle={Proceedings of the IEEE/CVF Conference on Computer Vision and Pattern Recognition (CVPR) Workshops},  
year = {2026} 
}

@inproceedings{ntire26reflection, 
title={{    NTIRE 2026 Challenge on Single Image Reflection Removal in the Wild: Datasets, Results, and Methods    }}, 
author={    Cai, Jie and  Yang, Kangning and  Li, Zhiyuan and  Vasluianu, Florin and  Timofte, Radu and others    },   
booktitle={Proceedings of the IEEE/CVF Conference on Computer Vision and Pattern Recognition (CVPR) Workshops},  
year = {2026} 
}

@inproceedings{ntire26anomalydet, 
title={{    NTIRE 2026  Challenge Report on Anomaly Detection of Face Enhancement for UGC Images    }}, 
author={    Zhong, Yan and   Ma,  Qiufang and  Wang, Zhen and  Jiang, Tingting and  Timofte, Radu and others    },   
booktitle={Proceedings of the IEEE/CVF Conference on Computer Vision and Pattern Recognition (CVPR) Workshops},  
year = {2026} 
}

@inproceedings{ntire26videosal, 
title={{    NTIRE 2026 Challenge on Video Saliency Prediction: Methods and Results    }}, 
author={    Moskalenko, Andrey and  Bryncev, Alexey and  Kosmynin, Ivan and  Shilovskaya, Kira and  Erofeev, Mikhail and  Vatolin, Dmitry and  Timofte, Radu and others    },   
booktitle={Proceedings of the IEEE/CVF Conference on Computer Vision and Pattern Recognition (CVPR) Workshops},  
year = {2026} 
}

@inproceedings{ntire26effsr, 
title={{    The Eleventh NTIRE 2026 Efficient Super-Resolution Challenge Report    }}, 
author={    Ren, Bin and  Guo, Hang and  Shu, Yan and  Ma, Jiaqi and  Cui, Ziteng and  Liu, Shuhong  and  Mei, Guofeng  and  Sun, Lei and  Wu, Zongwei and  Khan, Fahad Shahbaz and  Khan, Salman and  Timofte, Radu and  Li, Yawei and others    },   
booktitle={Proceedings of the IEEE/CVF Conference on Computer Vision and Pattern Recognition (CVPR) Workshops},  
year = {2026} 
}

@inproceedings{ntire26realx3d, 
title={{    3D Restoration and Reconstruction in Adverse Conditions: RealX3D Challenge Results    }}, 
author={    Liu, Shuhong and  Cui, Ziteng and  Bao, Chenyu and  Chu, Xuangeng and  Gu, Lin and  Ren, Bin and  Timofte, Radu and  Conde, Marcos V. and others    },   
booktitle={Proceedings of the IEEE/CVF Conference on Computer Vision and Pattern Recognition (CVPR) Workshops},  
year = {2026} 
}

@inproceedings{ntire26denoising, 
title={{    The Third Challenge on Image Denoising at NTIRE 2026: Methods and Results    }}, 
author={    Sun, Lei and  Guo, Hang and  Ren, Bin and  Su, Shaolin and  Wang, Xian and  Pani Paudel, Danda and  Van Gool, Luc and  Timofte, Radu and  Li, Yawei and others    },   
booktitle={Proceedings of the IEEE/CVF Conference on Computer Vision and Pattern Recognition (CVPR) Workshops},  
year = {2026} 
}

@inproceedings{ntire26aberration, 
title={{    NTIRE 2026 The First Challenge on Blind Computational Aberration Correction: Methods and Results    }}, 
author={    Sun, Lei and  Qian, Xiaolong and  Jiang, Qi and  Wang, Xian and  Gao, Yao and  Yang, Kailun and  Wang, Kaiwei and  Timofte, Radu and  Pani Paudel, Danda and  Van Gool, Luc and others    },   
booktitle={Proceedings of the IEEE/CVF Conference on Computer Vision and Pattern Recognition (CVPR) Workshops},  
year = {2026} 
}

@inproceedings{ntire26eventblurr, 
title={{    The Second Challenge on Event-Based Image Deblurring at NTIRE 2026: Methods and Results    }}, 
author={    Sun, Lei and  Li, Weilun and  Wang, Xian and  Li, Zhendong and  Shi, Letian and  Xu, Dannong and  Zhang, Deheng and  Hu, Mengshun and  Guo, Shuang and  Su, Shaolin and  Timofte, Radu and  Pani Paudel, Danda and  Van Gool, Luc and others    },   
booktitle={Proceedings of the IEEE/CVF Conference on Computer Vision and Pattern Recognition (CVPR) Workshops},  
year = {2026} 
}

@inproceedings{ntire26bursthdr, 
title={{    NTIRE 2026 Challenge on Efficient Burst HDR and Restoration: Datasets, Methods, and Results    }}, 
author={    Park, Hyunhee and  Park, Eunpil and  Lee, Sangmin and  Timofte, Radu and others    },   
booktitle={Proceedings of the IEEE/CVF Conference on Computer Vision and Pattern Recognition (CVPR) Workshops},  
year = {2026} 
}

@inproceedings{ntire26twilight, 
title={{    NTIRE 2026 Low-light Enhancement: Twilight Cowboy Challenge    }}, 
author={    Khalin, Aleksei and  Ershov, Egor and  Panshin, Artem and  Korchagin, Sergey and  Lobarev, Georgiy and  Terekhin, Arseniy and  Dorogova, Sofiia and  Shamsutdinov, Amir and  Mamedov, Yasin and  Khalfin, Bakhtiyar and  Sheludko, Bogdan and  Zilyaev, Emil and  Banić, Nikola and  Perevozchikov, Georgy and  Timofte, Radu and others    },   
booktitle={Proceedings of the IEEE/CVF Conference on Computer Vision and Pattern Recognition (CVPR) Workshops},  
year = {2026} 
}
